\documentclass[10pt,twocolumn,letterpaper]{article}
\usepackage{multirow}
\usepackage{iccv}
\usepackage{times}
\usepackage{epsfig}
\usepackage{graphicx}
\usepackage{amsmath}
\usepackage{amssymb}

\usepackage{booktabs}
\usepackage{mathtools}
\usepackage{enumerate}
\usepackage{adjustbox}
\usepackage{comment}
\usepackage{xcolor}
\usepackage{nicefrac}
\usepackage{authblk}
\newcommand{\R}{\mathbb{R}}
\newcommand\norm[1]{\left\lVert#1\right\rVert}

\usepackage[breaklinks=true,bookmarks=false]{hyperref}

\definecolor{DarkGreen}{rgb}{0.0, 0.7, 0.5}
\newcommand{\sara}[1]{\textcolor{olive}{[\textbf{Sara:} #1]} }

\usepackage[capitalize]{cleveref}
\crefname{section}{Sec.}{Secs.}
\Crefname{section}{Section}{Sections}
\Crefname{table}{Table}{Tables}
\crefname{table}{Tab.}{Tabs.}

\makeatletter
\DeclareRobustCommand\onedot{\futurelet\@let@token\@onedot}
\def\@onedot{\ifx\@let@token.\else.\null\fi\xspace}

\newcommand{\Table}[1]{Table~\ref{tab:#1}}
\newcommand{\Figure}[1]{Figure~\ref{fig:#1}}

\newcommand{\SM}{\textbf{Appendix}\xspace}

\newcommand{\best}[1]{\textbf{#1}}
\newcommand{\methodname}{{\fontfamily{cmss}\selectfont Re\nobreakdashes-ReND}\xspace}
\renewcommand\AB@affilsepx{, \protect\Affilfont}

\def\ie{\emph{i.e}\onedot}

\usepackage{array}
\newcommand{\thickhline}{%
    \noalign {\ifnum 0=`}\fi \hrule height 1pt
    \futurelet \reserved@a \@xhline
}
\newcolumntype{"}{@{\hskip\tabcolsep\vrule width 1pt\hskip\tabcolsep}}

\iccvfinalcopy 

\def\iccvPaperID{****} 

\ificcvfinal\fi

\begin{document}

\title{Re-ReND: Real-time Rendering of NeRFs across Devices}

\author[1]{Sara Rojas}
\author[1]{Jesus Zarzar}
\author[1]{Juan C. Pérez}
\author[2]{\authorcr Artsiom Sanakoyeu}
\author[2]{Ali Thabet}
\author[2]{Albert Pumarola}
\author[1]{Bernard Ghanem}
\affil[1]{KAUST} 
\affil[2]{Meta Research}


\maketitle
\ificcvfinal\thispagestyle{empty}\fi

\begin{abstract}
This paper proposes a novel approach for rendering a pre-trained Neural Radiance Field (NeRF) in real-time on resource-constrained devices.
We introduce \textbf{\methodname}, a method enabling \underline{Re}al-time \underline{Re}ndering of \underline{N}eRFs across \underline{D}evices. 
\methodname is designed to achieve real-time performance by converting the NeRF into a representation that can be efficiently processed by standard graphics pipelines.
The proposed method distills the NeRF by extracting the learned density into a mesh, while the learned color information is factorized into a set of matrices that represent the scene's light field.
Factorization implies the field is queried via inexpensive MLP-free matrix multiplications, while using a light field allows rendering a pixel by querying the field a single time---as opposed to hundreds of queries when employing a radiance field.
Since the proposed representation can be implemented using a fragment shader, it can be directly integrated with standard rasterization frameworks.
Our flexible implementation can render a NeRF in real-time with low memory requirements and on a wide range of resource-constrained devices, including mobiles and AR/VR headsets.
Notably, we find that \methodname can achieve over a 2.6-fold increase in rendering speed versus the state-of-the-art without perceptible losses in quality.

\end{abstract}

\vspace{-.5cm}
\section{Introduction}
\label{sec:intro}
Neural Radiance Fields (NeRFs)~\cite{mildenhall2021nerf} have revolutionized the field of novel view synthesis, as demonstrated by their impressive capacity to reconstruct complex objects and scenes with remarkable detail~\cite{barron2021mipnerf, tewari2020state}. 
The impressive performance of NeRFs casts them as a decisive tool for capturing and representing 3D objects and scenes.
As a result, NeRFs hold great promise for countless practical applications, including video games, movies, and Augmented/Virtual Reality (AR/VR).

However, the impressive performance of NeRFs comes with significant computational costs when rendering novel views.
This slow rendering is mainly due to two limiting properties of NeRFs, which dramatically increase their computational requirements.
Firstly, they use a volumetric representation to model scenes~\cite{drebin1988volume}, implying that rendering a single pixel requires hundreds of queries in space.
Secondly, they leverage a Multilayer Perceptron~(MLP) to model radiance and density across space, meaning that each of those spatial queries involves evaluating an expensive MLP.
These properties make it challenging to render NeRFs in real-time on resource-constrained devices, which hinders their practical deployment.

\begin{figure} 
 \centering
 \includegraphics[width=\columnwidth]{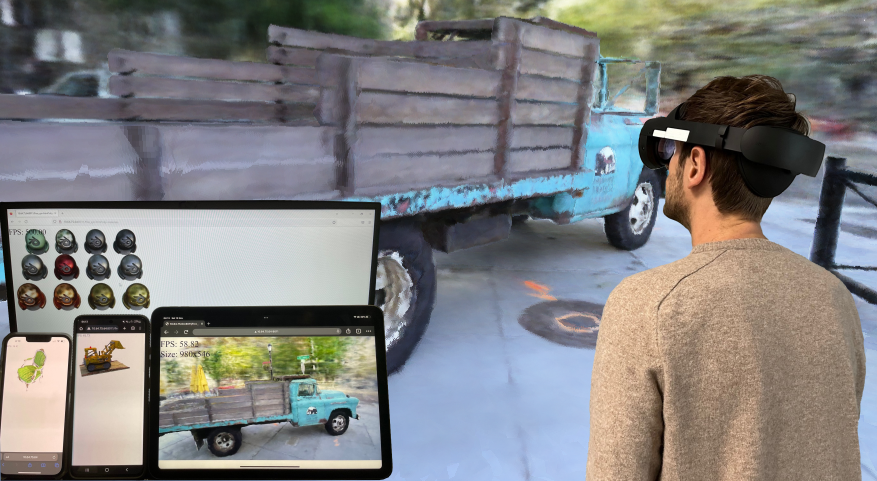}
 \caption{
     \textbf{Rendering NeRFs in real-time on resource-constrained devices, such as AR/VR headsets and mobiles.}
     We present \methodname, a method for rendering a pre-trained NeRF in real-time on a variety of devices with constrained computational resources.
     \methodname preserves remarkable photo-metric quality even when rendering at over 1,013 FPS on a desktop browser, or at the capped 74 FPS of a VR headset. 
     Please, refer to the accompanying video for a demonstration of \methodname's capacities.
}
 \vspace{-0.6cm}
 \label{fig:pull}
\end{figure}

A plethora of research efforts target the efficiency of NeRFs.
A line of work focused on shortening training times~\cite{yu_and_fridovichkeil2021plenoxels, kangle2021dsnerf, mueller2022instant}, while another looked at accelerating rendering times~\cite{neff2021donerf, reiser2021kilonerf}.
Notably, some methods pre-tabulate a NeRF's output on a sparse grid, and achieve real-time rendering by leveraging powerful GPUs~\cite{yu2021plenoctrees, hedman2021snerg}.
However, these methods are still incompatible with widely-available graphics pipelines that enable rendering on resource-constrained devices via popular frameworks such as WebGL and OpenGL ES.
This incompatibility stems from the rendering approach inherited from NeRFs, as volume rendering demands ray marching, which is dramatically more expensive than mesh rasterization.

In this paper, we present \textbf{\methodname}, a method that enables \underline{Re}al-time \underline{Re}ndering of \underline{N}eRFs across \underline{D}evices, including resource-constrained devices such as VR headsets and mobile phones.
To attain this end goal, we define and achieve three objectives: 
\textit{(i)}~enabling compatibility with widely-available graphics pipelines,
\textit{(ii)}~obtaining ray color with a single query, and
\textit{(iii)}~avoiding MLP evaluations for such queries.
Given a pre-trained NeRF as input, \methodname renders it in real-time by transforming the knowledge learned by the NeRF into an alternative representation. 
In particular, \methodname distills the NeRF by extracting the learned density into a mesh, and the learned color information into a set of matrices that efficiently factorize the scene's light field.
\methodname is thus capable of rendering a NeRF in real-time, making it deployable on a wide range of devices. 
When rendering challenging unbounded real scenes, our method achieves over 2.6-fold speed improvements above the state-of-the-art, while maintaining comparable quality. 
Furthermore, given that our rendering entirely disposes of MLPs, we can easily deploy it on other constrained devices, like VR headsets, where other methods cannot adapt. 
Please refer to Figure~\ref{fig:pull} for an overview of the capabilities of \methodname.

In summary, our contributions are threefold: 
    \textbf{(i)} We introduce \methodname, a method enabling real-time rendering of a pre-trained NeRF on resource-constrained devices.
    \methodname works on a wide variety of NeRFs, achieves remarkable rendering speeds at negligible costs to photo-metric quality, and is compatible with popular graphics pipelines available in common devices.
    \textbf{(ii)} \methodname enables fast rendering of a NeRF by transforming it into a representation with three fundamental qualities: 
    it resembles a graphics-friendly representation (\ie a mesh-texture package), 
    it obtains ray color via light fields (avoiding expensive volume rendering), and 
    it factorizes the light field computation as an MLP-free matrix multiplication.
    \textbf{(iii)} We conduct a comprehensive empirical study of \methodname across resource-constrained devices.
    Our results find remarkable boosts in rendering time that come at insignificant costs to image quality.
    Notably, we find \methodname boosts rendering speeds by $2.6\times$ at low memory expenses in challenging real scenes, enabling real-time rendering on various devices. 

Striving for reproducibility, we provide our implementation of \methodname, written in PyTorch~\cite{NEURIPS2019_9015}, in the \SM.
\vspace{-0.6cm}
\begin{figure*}[t!]
    \centering
    \includegraphics[width=\textwidth]{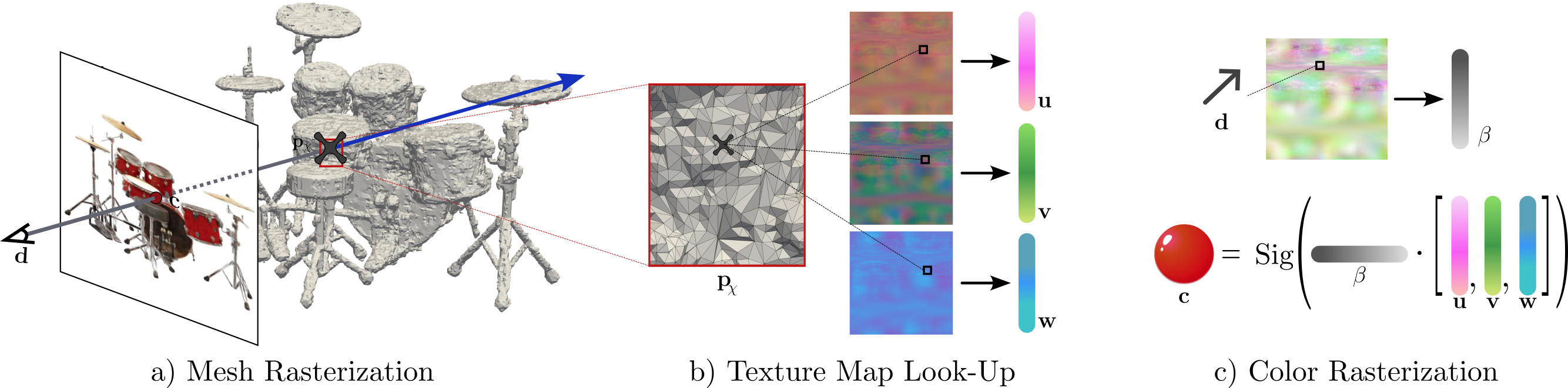}
    \vspace{-0.4cm}
    \caption{
    \textbf{Rendering a NeRF using \methodname.} 
    Given a pre-trained NeRF, \methodname renders it in real time by leveraging a mesh and light field embedding maps ($\mathbf{u}, \mathbf{v}, \mathbf{w}$ and $\boldsymbol\beta$) which are compatible with a standard rasterization pipeline with programmable shaders.
    Rendering occurs in three steps:
    \textbf{a)} The mesh gets rasterized, and its vertices go through a regular vertex shader. 
    \textbf{b)} The light field embedding maps are indexed (with the $uv$ coordinates from the rasterization) to obtain values for the position-dependent ($\mathbf{u}, \mathbf{v}$ and $\mathbf{w}$) and direction-dependent ($\boldsymbol\beta$) embeddings. 
    \textbf{c)} Pixel colors are obtained by combining $\mathbf{u}, \mathbf{v}, \mathbf{w}$ and $\boldsymbol\beta$ in a custom fragment shader implementing an inexpensive dot product. 
    Since \methodname uses light fields, alpha compositing is unnecessary, and so we set alpha values to 1. 
    Note that our method entirely disposes of MLPs at render time, and thus enjoys substantial boosts in rendering speed.
    }
    \vspace{-0.4cm}
    \label{fig:Rendering}
\end{figure*}

\section{Related Work}\label{sec:relatedwork}
\vspace{-0.5em}
\vspace{2pt}\noindent\textbf{Light Field Representations.}
Light fields~\cite{levoy1996light} represent a scene as integrated radiance along rays, \ie they are the integral of radiance fields.
Unlike radiance fields, light fields directly map ray origin and direction to color, allowing for a single query to compute a pixel's color instead of multiple queries.
Innovative works have aimed at parameterizing light fields with neural networks~\cite{sitzmann2021light, suhail2022light, attal2022learning, feng2021signet, r2l}.
However, despite their desirable properties, light field representations have been found to be challenging to learn~\cite{lassner2021pulsar}.
This difficulty mainly stems from 
\textit{(i)}~their reductive formulation of appearance requiring compensation with clever parameterizations~\cite{sitzmann2021light, feng2021signet} or larger and data-hungry architectures~\cite{r2l}, and 
\textit{(ii)}~their indirect use of geometry that calls for injecting geometric priors elsewhere in the pipeline~\cite{suhail2022light, attal2022learning}.
Our proposed \methodname renders a pre-trained NeRF in real time by distilling the NeRF's color knowledge into a light field, thus leveraging the inherent one-query-per-pixel virtue of these fields.
In particular, our light field formulation is similar to~\cite{wood2000surface}, whereby
the field generates the color at the point of intersection between the camera ray and the scene's geometry. 
We further accelerate rendering by factorizing the light field into matrices, and enabling compatibility with graphics pipelines by storing such matrices into texture~map-like arrays.

\vspace{2pt}\noindent\textbf{Factorizing Neural Fields.}
NeRFs~\cite{mildenhall2021nerf} use an MLP to map a position $\mathbf{p} \in \mathbb{R}^3$ and a view direction $\mathbf{d} \in \mathbb{R}^2$ to color and density.
Various works have studied the complex interplay between position and view direction that happens inside NeRFs~\cite{hedman2021snerg, garbin2021fastnerf, yu2021plenoctrees, yu_and_fridovichkeil2021plenoxels, Wizadwongsa2021NeX}.
Notably, Garbin~\etal~\cite{garbin2021fastnerf} highlight how caching a NeRF's output is precluded by the dependence of color on \textit{both} $\mathbf{p}$ and $\mathbf{d}$.
Thus, the authors propose to factorize color into two independent functions (separately for $\mathbf{p}$ and $\mathbf{d}$), such that the inner product of their outputs generates color.
This factorization allows for caching the NeRF's output, and thus effectively disposes of MLPs for rendering.
Our work enables real-time rendering by combining this factorization approach with the color formulation of light fields.
Specifically, we leverage the NeRF-factorization method of~\cite{garbin2021fastnerf} to factorize a Neural \textit{Light} Field.
This combination allows \methodname to enjoy both the factorization's MLP-free rendering, and the one-query-per-pixel virtue of light fields.

\vspace{2pt}\noindent\textbf{Rendering of Neural Fields.}
Neural fields achieve impressive photo-realistic quality at undesirably large computational costs~\cite{mildenhall2021nerf}.
Various works have addressed training costs~\cite{mueller2022instant, yu_and_fridovichkeil2021plenoxels, Chen2022ECCV, SunSC22}, while others focused on rendering.
For rendering, advances in differentiable and fast rendering~\cite{kato2020differentiable, Niemeyer2020CVPR, cole2021differentiable} have enabled novel advances and applications~\cite{zhang2021nerfactor}.
While several approaches achieve real-time rendering on power-intensive setups~\cite{garbin2021fastnerf, yu2021plenoctrees, reiser2021kilonerf, lindell2021autoint, rebain2021derf} with access to GPUs, we focus on enabling real-time rendering on resource-constrained devices such as AR/VR headsets and mobiles.
Rendering can be accelerated by exploiting graphics pipelines offered by the hardware of these devices. 
Both SNeRG~\cite{hedman2021snerg} and PlenOctrees~\cite{yu2021plenoctrees} leveraged optimized graphics routines for in-browser rendering, but did not exploit efficient mesh-oriented pipelines.
These pipelines are difficult to exploit in NeRF's formulation, since the volume rendering~\cite{drebin1988volume} nature of NeRFs is incompatible with the polygon-oriented paradigm of mesh rasterization.
In this work, we accelerate the rendering of NeRFs by transforming their learned knowledge into a mesh-friendly representation on which efficient mesh-rasterization pipelines can operate. 
In particular, we represent a scene as a mesh whose texture maps are filled with ``light field embeddings'', which result from factorizing the scene's light field.
Concurrently with our work, Chen~\etal~\cite{chen2022mobilenerf} propose MobileNeRF, which also exploits mesh-rasterization pipelines for fast rendering on devices. 
While MobileNeRF uses surface-based neural fields, \methodname uses light fields at the surface of objects.
Furthermore, our method acts as a framework for \textit{transforming} a pre-trained NeRF to an MLP-free and rasterization-friendly representation that is capable of real-time rendering across devices. 

\section{Problem Formulation}
\label{sec:formulation}
Given a pre-trained NeRF, we aim at rendering it in real time on a variety of resource-constrained devices.
Our goal is to transform said NeRF into a representation that runs on standard mesh-rasterization pipelines.
A NeRF is a function $R$ (parameterized by an MLP), which maps from a 3D location $\mathbf{p}$ and a 2D viewing direction $\mathbf{d}$ to an RGB emitted color $\mathbf{c}$ and volume density $\sigma$.
Formally, a NeRF implements $R~:~(\mathbf{p},\mathbf{d})~\mapsto~(\mathbf{c},\sigma)$.
To render a pixel, NeRF follows the classical volume rendering formulation.
In this formulation, the color assigned to a pixel with associated camera ray $\mathbf{r}(t) = \mathbf{o} + t\:\mathbf{d}$ is given by:
\begin{equation}\label{eq:color}
C(\mathbf{r}) = \int_{t_n}^{t_f}T(t)\:\sigma\left(\mathbf{r}(t)\right)\:\mathbf{c}(\mathbf{r}(t), \mathbf{d})\:\mathrm{d}t,
\end{equation}
where $T(t) = \exp\left(-\int_{t_n}^{t}\sigma\left(\mathbf{r}(s)\right)\:\mathrm{d}s\right)$ represents the transmittance accumulated along $\mathbf{r}$.

To reach our goal of real-time rendering of NeRFs in resource-constrained devices,
we must address the following three limitations of NeRFs:
\textit{(i)}~their implicit nature 
conflicts with the explicit representations demanded by mesh-rasterization pipelines, which request a mesh $\chi$ and a texture map $\mathbf{M}$, 
\textit{(ii)}~they compute color via Eq.~\eqref{eq:color}, thus requiring expensive numerical estimation of the integral, and 
\textit{(iii)}~computing the individual integrands for this estimate requires evaluating the MLP that parameterizes $R$.

We seek a method that transforms $R$'s knowledge into an alternative representation that circumvents these three limitations of NeRFs. 
That is, we require the method to output a rasterization-friendly representation,
to compute pixel color with a single query, 
and to entirely dispose of MLPs for modeling view-dependent effects.
Moreover, we require the method to effectively work on a wide variety of NeRFs.
\vspace{-0.6cm}

\begin{figure*}[t!]
    \includegraphics[width=\linewidth]{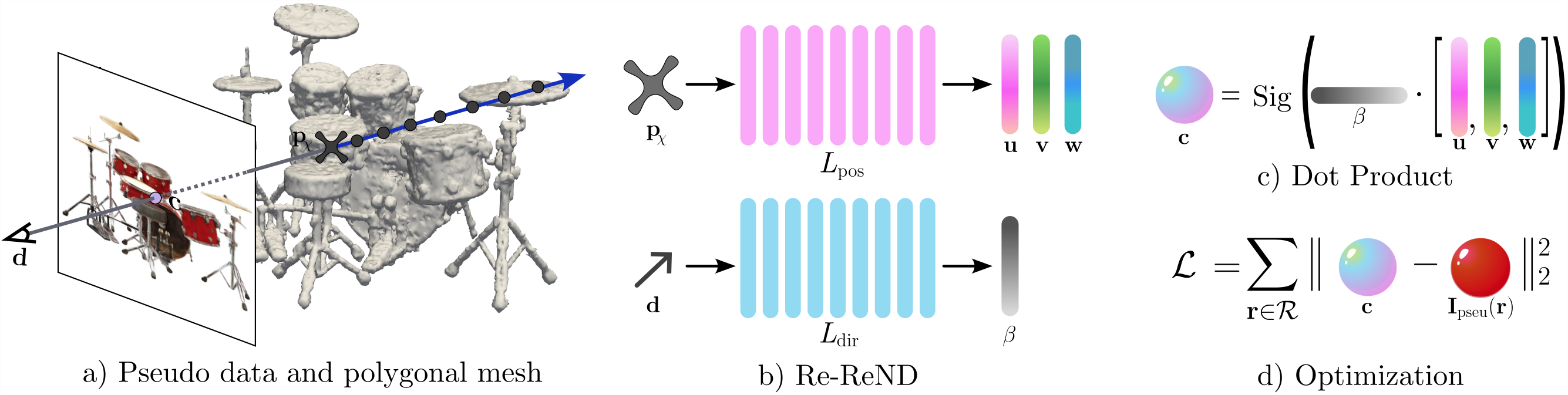}
    \vspace{-0.4cm}
    \caption{
    \textbf{Training \methodname.} 
    \textbf{a)} Starting with a pre-trained NeRF, we extract a polygonal mesh and generate pseudo-images to train our factorized NeLF. 
    These images are generated by rendering the NeRF from various points of view.
    \textbf{b)} Our factorized NeLF consists of two MLPs ($L_\text{pos}$ and $L_\text{dir}$) that \textit{independently} compute position and direction embeddings. 
    $L_\text{pos}$ outputs position-dependent deep radiance embeddings $(\mathbf{u}, \mathbf{v}, \mathbf{w})$ for points \textit{on} the mesh's surface, while $L_\text{dir}$ outputs direction-dependent embeddings ($\boldsymbol\beta$). 
    Since this representation is amenable to ``baking'',~\ie~pre-computing and storing outputs, it allows us to dispose of MLPs for deployment. 
    \textbf{c)}~Under this formulation, computing pixel color amounts to a dot product of $(\mathbf{u}, \mathbf{v}, \mathbf{w})$ and $\boldsymbol\beta$. 
    \textbf{d)}~We optimize our framework with an MSE reconstruction loss w.r.t. to the pseudo-images we extracted from the pre-trained NeRF.
    }
    \vspace{-0.5cm}
    \label{fig:Training}
\end{figure*}
\section{Method: \methodname}
\label{sec:methodology}

We now introduce \methodname, a method enabling real-time rendering of a pre-trained NeRF on resource-constrained devices.
\methodname receives a pre-trained NeRF $R$ as input and transfers its knowledge to an alternative representation $\mathcal{S}$ that is 
fast and inexpensive to render.
The representation generated by \methodname achieves fast rendering by 
\textit{(i)}~representing the learnt scene as a mesh whose texture maps store embeddings, 
\textit{(ii)}~obtaining ray color via light fields~(instead of radiance fields required by expensive volume rendering), and 
\textit{(iii)}~factorizing the light field computation as an MLP-free matrix multiplication.
Please refer to Figure~\ref{fig:Rendering} for an overview of \methodname.

In essence, \methodname transforms a pre-trained NeRF $R$ into a rasterization-friendly representation $\mathcal{S} = \{\chi,\mathcal{M}\}$.
This representation is composed of a mesh $\chi$, 
and a set of four texture-map-like matrices $\mathcal{M} = \{ \mathbf{M}_\mathbf{u}, \mathbf{M}_\mathbf{v}, \mathbf{M}_\mathbf{w}, \mathbf{M}_{\boldsymbol\beta} \}$.
Each matrix $\mathbf{M}_i$ 
stores representations we dub ``light field embeddings'', where position-dependent effects are modeled by the first three matrices, while the fourth one accounts for view-dependent effects.
While obtaining the mesh $\chi$ is straightforward from the densities generated by $R$, 
constructing $\mathcal{M}$ is challenging.

In the next two sections, we describe how we obtain~$\mathcal{M}$.
In particular, Section~\ref{sec:nelfs} describes how we approximate $R$ with a Factorized Neural Light Field (NeLF), and Section~\ref{sec:fnelfs2rere} explains how we extract $\mathcal{M}$ from this NeLF.

\subsection{Representing $R$ with Factorized NeLFs}\label{sec:nelfs}
We construct $\mathcal{M}$ by using the radiance field $R$ as supervision to train a \textit{Factorized} Neural Light Field~(NeLF).
Factorized NeLFs are light fields whose computation can be expressed as inexpensive matrix multiplications.
We first explain the formulation for constructing a NeLF, and then the formulation for their factorization. 

\vspace{2pt}\noindent\textbf{NeLF.}
We extract the color information learned by the NeRF $R$ into a NeLF $L$.
Light fields are the integral of radiance fields: while radiance fields reconstruct the individual integrands of Eq.~\eqref{eq:color} along the ray, light fields directly estimate the integral's value, \ie~the ray's color.
Thus, light fields are more suitable for our purposes, as they directly yield $C(\mathbf{r})$ instead of computing the integral of Eq.~\eqref{eq:color}.

Formally, given a position $\mathbf{p}$ and direction $\mathbf{d}$, the light field models the integrated radiance along the ray with origin at $\mathbf{p}$ and direction $\mathbf{d}$, \ie
$C(\mathbf{p} + t\:\mathbf{d}),\:t\in[0,+\infty)$.
If the scene's geometry is known, empty space can be skipped by considering a ray whose origin is the collision point of the ray and the scene's geometry.
We use our mesh $\chi$ as an estimate to this geometry, and define $\mathbf{p}^{\mathbf{r}}_\chi \coloneqq \mathbf{r} \wedge \chi$ as the first intersection point between ray $\mathbf{r}$ and mesh $\chi$.
The NeLF $L$ thus must approximate $C$, that is,
\vspace{-0.2cm}
\begin{equation}\label{eq:red_light_field}
\vspace{-0.2cm}
    L(\mathbf{p}^{\mathbf{r}}_\chi,\mathbf{d}) \approx 
    C(\mathbf{p} + t\:\mathbf{d}),\:t\in[0,+\infty).
\end{equation} 

Inspired by~\cite{r2l}, and using the formulation of Eq.~\eqref{eq:red_light_field}, we can distill a NeRF $R$ into a NeLF $L$.
Specifically, we can use $R$ to compute the right hand side of this equation, and use these values as supervision to learn $L$. 
Thus, 
the NeLF is tasked with predicting the color of ray $\mathbf{r}$ when evaluated at the collision point $\mathbf{p}^{\mathbf{r}}_\chi$.
In practice, we sample camera views (and their associated set of rays $\mathcal{R}$) and use $R$ to compute a set of pseudo-images~$\mathcal{I}_{\text{pseu}}$. 
Furthermore, we use $\chi$ to pre-compute the collision points between the rays in $\mathcal{R}$ and the scene.
We then evaluate $L$ at these points, and task it with predicting the colors given by~$\mathcal{I}_{\text{pseu}}$.

\vspace{2pt}\noindent\textbf{Factorized NeLFs.}
For a given camera ray, the NeLF $L$ predicts ray color when evaluated at the collision point of the camera ray and the scene.
Because of the light field formulation, this approach can compute pixel colors with a single MLP query. 
However,~$L$'s evaluations are still expensive due to its MLP parameterization.
Hence, we propose to rather learn a Factorized NeLF.
The architecture of a Factorized NeLF is amenable to ``baking'',~\ie~pre-computing and storing network outputs, to dispose of MLPs at rendering time.
We next describe Factorized NeLFs in detail.

Recall that our NeLF is a function $L~:~(\mathbf{p}^{\mathbf{r}}_\chi,\mathbf{d})\mapsto\mathbf{c}$, parameterized by an MLP.
This function maps a point~(on the mesh)~$\mathbf{p}^{\mathbf{r}}_\chi \in \R^3$ and a ray direction~$\mathbf{d} \in \R^2$ to an RGB color $\mathbf{c} \in \R^3$.
A \textit{Factorized} NeLF $L_F$ shares the same signature with $L$, but internally processes $\mathbf{p}^{\mathbf{r}}_\chi$ and $\mathbf{d}$ with two independent MLPs whose outputs produce color via an inexpensive matrix multiplication.
We define these underlying direction- and position-dependent MLPs~\cite{garbin2021fastnerf} as
\begin{equation}\label{eq:dir-pos}
\begin{aligned}
    L_{\text{pos}}: &\:\mathbf{p}^{\mathbf{r}}_\chi \mapsto [\mathbf{u}, \mathbf{v}, \mathbf{w}]  \in \R^{D\times 3}, \\
    L_{\text{dir}}: &\:\mathbf{d}                   \mapsto \boldsymbol\beta \in \R^{D}.
\end{aligned}
\end{equation}
We refer to the vectors 
$\mathbf{u}$, $\mathbf{v}$, $\mathbf{w}$, 
and 
$\boldsymbol\beta$
as \textit{light field embeddings}.
Our Factorized NeLF $L_F$ is thus defined as
\begin{equation} \label{eq:factorized-nelf}
\begin{aligned}
    L_F(\mathbf{p}^{\mathbf{r}}_\chi, \mathbf{d}) &= \mathrm{Sig}\left(L_{\text{pos}}(\mathbf{p}_\chi)^\top L_{\text{dir}}(\mathbf{d})\right) \\
    &= \mathrm{Sig}\left([\mathbf{u}, \mathbf{v}, \mathbf{w}]^\top\boldsymbol\beta\right),
\end{aligned}
\end{equation}
where $\mathrm{Sig}$ is the sigmoid function.
Note, in Eq.~\eqref{eq:factorized-nelf}, the position-dependent embeddings ($\mathbf{u}$, $\mathbf{v}$, $\mathbf{w}$) are weighed by the direction-dependent embeddings $\boldsymbol\beta$.
This formulation enables accelerated rendering via baking: $L_{\text{pos}}$ and $L_{\text{dir}}$ outputs are pre-computed and stored, so that $L_F$ can be approximated at test time with an inexpensive MLP-free operation.

\vspace{2pt}\noindent\textbf{Training Factorized NeLFs.}
We train $L_F$ to predict the colors given by the set of pseudo-images~$\mathcal{I}_{\text{pseu}}$ when evaluated at the point of collision with the scene.
Formally, 
we train~$L_F$ to minimize the photo-metric loss
\begin{equation}\label{eq:fnelf_loss}
    \mathcal{L} = \sum_{\mathbf{r} \in \mathcal{R}}\norm{L_F\left(\mathbf{p}^{\mathbf{r}}_\chi, \mathbf{d}\right) - \mathcal{I}_{\text{pseu}}(\mathbf{r})}_2^2,
\end{equation}
where $\mathcal{R}$ is the set of rays with which the set of pseudo-images $\mathcal{I}_{\text{pseu}}$ 
was computed, and $\mathcal{I}_{\text{pseu}}(\mathbf{r})$ further denotes the color assigned to ray $\mathbf{r}$ in the pseudo-images.

\subsection{Factorized NeLFs to Real-time Rendering}\label{sec:fnelfs2rere}
We now use our Factorized NeLF to build $\mathcal{M}$, the missing piece in our a rasterization-friendly representation $\mathcal{S} = \{\chi,\mathcal{M}\}$.
Recall $\mathcal{M} = \{ \mathbf{M}_\mathbf{u}, \mathbf{M}_\mathbf{v}, \mathbf{M}_\mathbf{w}, \mathbf{M}_{\boldsymbol\beta} \}$ is a set of four matrices, 
where the first three matrices model position-dependent effects, and the fourth one 
view-dependent effects.
Here we describe how we leverage our Factorized NeLF to construct $\mathcal{M}$, 
and then how we integrate this representation with mesh-rasterization pipelines.

\vspace{2pt}\noindent\textbf{Constructing $\mathcal{M}$.}
The Factorized NeLF formulation from Eq.~\eqref{eq:factorized-nelf} is amenable to baking,~\ie~pre-computation and storage.
Namely, if the outputs of $L_{\text{pos}}$ and $L_{\text{dir}}$ are baked, computing a ray's color is reduced to an inexpensive MLP-free operation.
To enjoy this MLP-free property when rendering with a mesh-rasterization pipeline, we use $\mathcal{M}$ to store the baked outputs.
Specifically, we pre-compute light field embeddings at various inputs and store them in the corresponding matrices $\mathbf{M}_\mathbf{u}$, $\mathbf{M}_\mathbf{v}$, $\mathbf{M}_\mathbf{w}$ and $\mathbf{M}_{\boldsymbol\beta}$.
For the position-dependent embeddings, we traverse $\chi$'s faces, extract $N_\text{pos}$ points in each face's normalized $uv$ coordinates, 
and then evaluate $L_{\text{pos}}$ at such positions.
Analogously, for the direction-dependent embeddings, 
we  evaluate $L_{\text{dir}}$ at independently sampled $N_\text{dir}$ azimuth and elevation angles.

Once constructed $\mathcal{M}$, we are ready to integrate our representation $\mathcal{S}$ with the rasterization pipeline.

\vspace{2pt}\noindent\textbf{Integration with mesh rasterization.}
We render our representation $\mathcal{S} = \{\chi,\mathcal{M}\}$ on a standard graphics pipeline, 
enabling real-time rendering across devices, from mobiles all the way to VR headsets.
Note that our scene representation is compatible with standard graphics pipelines: the required mesh is compatible with the $\chi$ mesh extracted from $R$, and the texture are our four matrices in $\mathcal{M}$.
In particular, 
as common in production rendering~\cite{burley2008ptex}, 
we store the position-dependent info ($\mathbf{M}_\mathbf{u}$, $\mathbf{M}_\mathbf{v}$ and $\mathbf{M}_\mathbf{w}$) into a texture map-like with per-face textures, and the direction-dependent info ($\mathbf{M}_{\boldsymbol\beta}$) in a texture map sorted by its corresponding sampling angles.
We then quantize all baked outputs, and export them as PNG files. 
Finally, we deploy our rendering pipeline within a fragment shader for compatibility with standard rasterization frameworks~(please refer to Figure~\ref{fig:Training}).

\section{Experiments}\label{sec:experiments}
\vspace{-0.2cm}
\begin{table}[t!]
    \footnotesize
    \centering
    \setlength{\tabcolsep}{1.5pt}
    \begin{tabular}{c"c"c"c"c}
        \thickhline
        Device         & Type    & OS              & GPU                     & Browser \\ 
        \thickhline
        Samsung S21    & Phone   & Android 13      & Mali G78                & Chrome  \\
        Motorola G9    & Phone   & Android 11      & Adreno 610              & Chrome    \\
        Galaxy S6      & Tablet  & Android 13       & Mali G72 MP3               & Firefox   \\
        Dell           & Laptop  & Windows 10      & Integrated GPU           & Firefox  \\
        Gaming Lap.    & Laptop  & Windows 10      & NVIDIA GF RTX 2060      & Firefox   \\
        Desktop             & Desktop & Ubuntu 18.04    & NVIDIA GF RTX 3090      & Chrome \\
        Meta Quest P. & Headset & Oculus & Adreno 650 & $-$ \\
        \thickhline
    \end{tabular}
    \caption{\textbf{Testing Devices.}
    We compare \methodname against other methods on a set of representative devices with a wide range of compute capabilities. 
    Our devices include low- to high-end mobile phones, tablets, laptops, desktop computers, and a VR headset. 
    }
    \vspace{-0.45cm}
    \label{tab:0_hdw_specs_tb}
\end{table}

\begin{table*}[t!]
    \footnotesize
    \centering
    \setlength{\tabcolsep}{2pt}
\begin{tabular}{l"l"c|c|c|c|c"c|c|c|c|c}\thickhline
       &             & \multicolumn{5}{c"}{Synthetic 360°}                  & \multicolumn{5}{c}{Unbounded 360°}\\
     Device    &             Method                                      & FPS        & GPU (MB)        & Disk (MB)      & PSNR  (dB)       & Mesh  (F/V)    & FPS    & GPU (MB)  & Disk (MB)          & PSNR (dB)   & Mesh (F/V) \\\thickhline
\multirow{3}{*}{Samsung} & M-NeRF  \cite{chen2022mobilenerf} & 41.7    &     570   & 125.8    &\best{30.9}& 494k/224k     & 22.8       &    943      &311.3& 15.6   & 494k/224k       \\
        & SNeRG \cite{hedman2021snerg}                    & $\dagger$         & 3,627         & \best{87.0} & 30.4      & N/A                &  $\dagger$      & 4,386   & 324.8     &      14.0      & N/A      \\
        & Ours                                     &\best{54.7}& \best{532}           & 199.0     & 29.0     & \best{205k/99k} &    \best{33.5}    & \best{816}  & \best{288.3}      & \best{17.9}         & \best{244k/117k} \\\thickhline
\multirow{7}{*}{Desktop} 
        & M-NeRF  \cite{chen2022mobilenerf}               & 124.3       & 7,350       & 125.7     & \best{30.9} & 494k/224k & 14.3 & 7,970  & 311.3 & 15.6 & 332k/543k        \\
        & SNeRG \cite{hedman2021snerg}                    & 49.1       & 3,963        &\best{87.0}& 30.4       & N/A             & 9.9  & 5,102 & 324.8        & 14.0 & N/A                \\
        & Ours                                      &\best{701.4}&\best{1,501}& 199.0      & 29.0      &\best{205k/99k}&\best{235.3}&\best{1,800}& \best{288.3} & 17.9 &\best{244k/117k}\\
        & Ours (Large)                                    & 329.6       & 2,964       & 800.7     & 30.1     &\best{205k/99k}& 118.6       & 3,445     & 1,142.9  &\best{18.0}&\best{244k/117k}\\\cline{2-12}
        & NeRF \cite{mildenhall2021nerf} &    7e-4      &   OOM         & 13.3       & 31.0          & N/A               & 7e-4      & OOM     & 13.3            & 19.0    & N/A                      \\
        & Mip-NeRF \cite{barron2021mipnerf}               & 1e-2       & OOM           & 7.4       & 33.1      & N/A               & $-$      & $-$     & $-$            & $-$    & $-$                    \\
        & NeRF++ \cite{nerf++}                            & $-$           & $-$           & $-$         & $-$          & $-$              &    8e-4    & OOM     & 29.0         &  20.1  & N/A                    \\  \thickhline

\end{tabular}
 \vspace{0.07cm}
\caption{\textbf{Quantitative Results.} We report rendering speed, GPU memory a disk requirements, PSNR, and number of mesh faces and vertices on a mobile device and a desktop computer for both datasets.
The last three rows include baseline methods, which do not run on mobile devices.
Rendering speed is measured at a resolution of 8K on a desktop and 800px on a mobile device (Samsung). 
GPU memory requirements are calculated for rendering a full frame. Methods unable to render an entire frame are denoted as Out of Memory (OOM).
Along with our base model, we report metrics on a variation ``Ours (Large)" with higher texel count (10) and larger embedding dimension (64).
We achieve a major speedup along with a higher quality rendering on the Unbounded 360 dataset and a slight degradation of quality in the Synthetic 360 dataset.
``$\dagger$" denotes the method was unable to run on device while ``$-$" denotes a missing implementation.
    }
    \vspace{-0.4cm}
    \label{tab:desktop}
\end{table*}

Next, we conduct an extensive study of \methodname performance on multiple scenes and devices.
\vspace{-0.5cm}
\paragraph{Datasets.}
We experiment on both synthetic and real data by using two standard datasets:
\vspace{-0.2cm}
\begin{itemize}
\item \textbf{Synthetic 360° dataset}~\cite{barron2022mip}, eight synthetic scenes with intricate geometries and non-Lambertian materials.
Each scene has 100 views for training and 200 views for testing, both at a resolution of $800$px$\times800$px.
\vspace{-0.2cm}
\item \textbf{Tanks and Temples (T\&T) dataset}~\cite{tandt}, an unbounded dataset consisting of hand-held 360° captures of four large-scale scenes. 
We use the dataset configuration of~\cite{nerf++}.
\vspace{-0.5cm}
\end{itemize}

\paragraph{Devices.}
Our main goal is to test our method on hardware-constrained devices such as mobile phones, tablets, and VR headsets. 
Nonetheless, for completeness, we also test \methodname on more powerful laptops and desktops. 
In total, we test \methodname on seven devices reported in \Table{0_hdw_specs_tb}. 
\vspace{-0.1cm}

\subsection{Implementation details} 
\vspace{-0.1cm}

\noindent\textbf{Pre-trained NeRFs.}
For the pre-trained NeRF models $R$, we use MipNeRF~\cite{barron2021mipnerf} (in the Synthetic 360° dataset), 
and NeRF++~\cite{nerf++} (in the T\&T dataset). 

\noindent\textbf{Meshing.}
Obtaining the mesh $\chi$ from a pre-trained NeRF requires distilling the learnt geometry. 
For the Synthetic 360° dataset, we run Marching Cubes~\cite{lorensen1987marching} on a density grid of side $K = 256$, except for the \textit{ficus} object, in which we use $K = 512$ to capture finer geometric details.
For the T\&T dataset, we use a grid with $K = 512$.
We remove small connected components resulting from noisy estimates, and decimate the meshes to around $400$k faces. 
Finally, we enclose unbounded scenes within a dome. 


\begin{figure}[t]
    \includegraphics[width=0.47\textwidth]{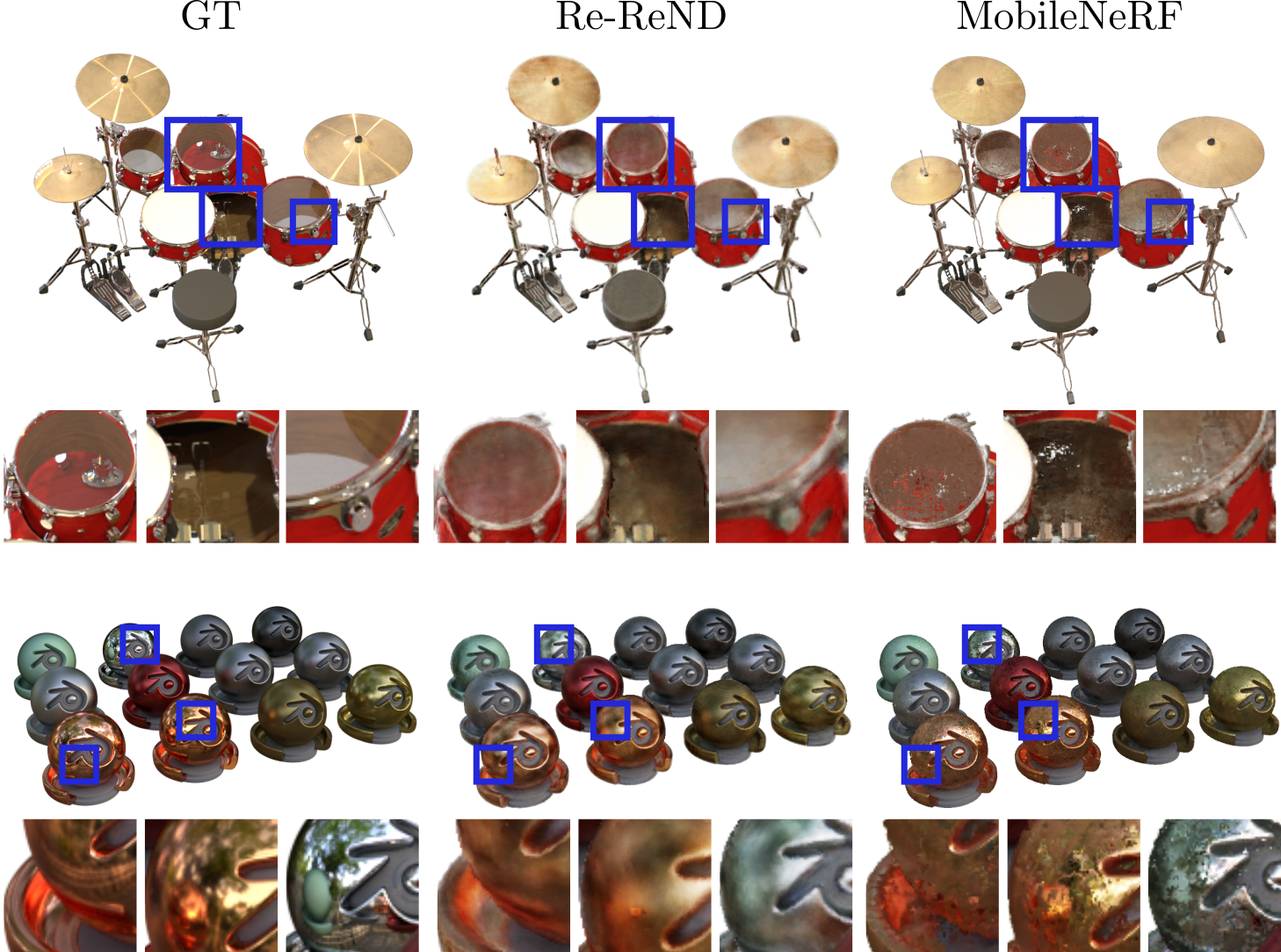}
    \caption{
    \textbf{Qualitative Results on Synthetic Scenes.}
    We show renderings of two synthetic scenes to compare \methodname and MobileNeRF~\cite{chen2022mobilenerf}. 
    In both scenes, we can see how \methodname accurately more reproduces light effects like shininess and reflections. Notice that MobileNeRF fails to model such scene producing holes and high frequency noise.
    }
    \vspace{-0.5cm}
    \label{fig:SyntheticQualitative}
\end{figure}

\noindent\textbf{Pseudo-Image Generation.}
Training the Factorized NeLF via Eq.~\eqref{eq:fnelf_loss} requires the set $\mathcal{I}_{\text{pseu}}$ of pseudo-images.
We obtain these images by using the NeRF to render $10$k images from random camera poses for each scene. 

\noindent\textbf{Factorized NeLF Training.}
We implement the position- and direction-dependent MLPs ($L_{\text{pos}}$ and $L_{\text{dir}}$) by following~\cite{r2l}, and thus employ intensive residual blocks~\cite{resnet} and deep architectures of $88$ layers. 
We train these MLPs with hard-ray sampling~\cite{r2l} and learning rate warm-up strategies.
We train with a batch size of $200$k rays for 2.5 days on one NVIDIA A100 GPU.

\noindent\textbf{Baking Light Field Embeddings.}
For baking $\boldsymbol\beta$, we uniformly sample elevation and azimuth angles in $[0$-$180^{\circ}]$ and $[0$-$360^{\circ}]$, respectively. 
We use $1024$ samples for synthetic scenes and $2048$ for real scenes.
When quantizing $\mathbf{u}$, $\mathbf{v}$, $\mathbf{w}$, and $\boldsymbol\beta$ for their storage as PNG files, we perform per-channel min-max normalization.
Unless otherwise stated, we use $18 = \lceil\nicefrac{6\times6}{2}\rceil$ texels per triangle face. 

\noindent\textbf{Rendering with Shaders.}
Since \methodname requires no MLP queries, we implement it in a simple fragment shader.
This implementation allows deploying \methodname across not only various devices, but also different graphics frameworks. 
Notably, this implementation allows us to deploy on a VR headset (Meta Quest Pro), which runs Unity shaders.
The shader computes color by combining the position- and direction-dependent embeddings according to Eq.~\eqref{eq:factorized-nelf}, where each embedding is queried from 
its corresponding texture map.
In turn, each texture map is a $2\times4$ grid stored in a $4$-channeled PNG image, 
thus fully accounting for our default embedding dimension $D = 32 = 2\times4\times4$.
We obtain the position embeddings by indexing $\mathbf{M}_{\mathbf{u},\mathbf{v},\mathbf{w}}$ with the fragment's $uv$ coordinates, while the direction embeddings are obtained by indexing $\mathbf{M}_{\boldsymbol\beta}$ with the azimuth and elevation angles.
Since texture values are 8-bit quantized, we map to the original range by reverting the per-channel min-max normalization. 
We provide a full implementation of our shaders in the \SM.


\begin{figure*}[t]
    \centering
    \includegraphics[width=\linewidth]{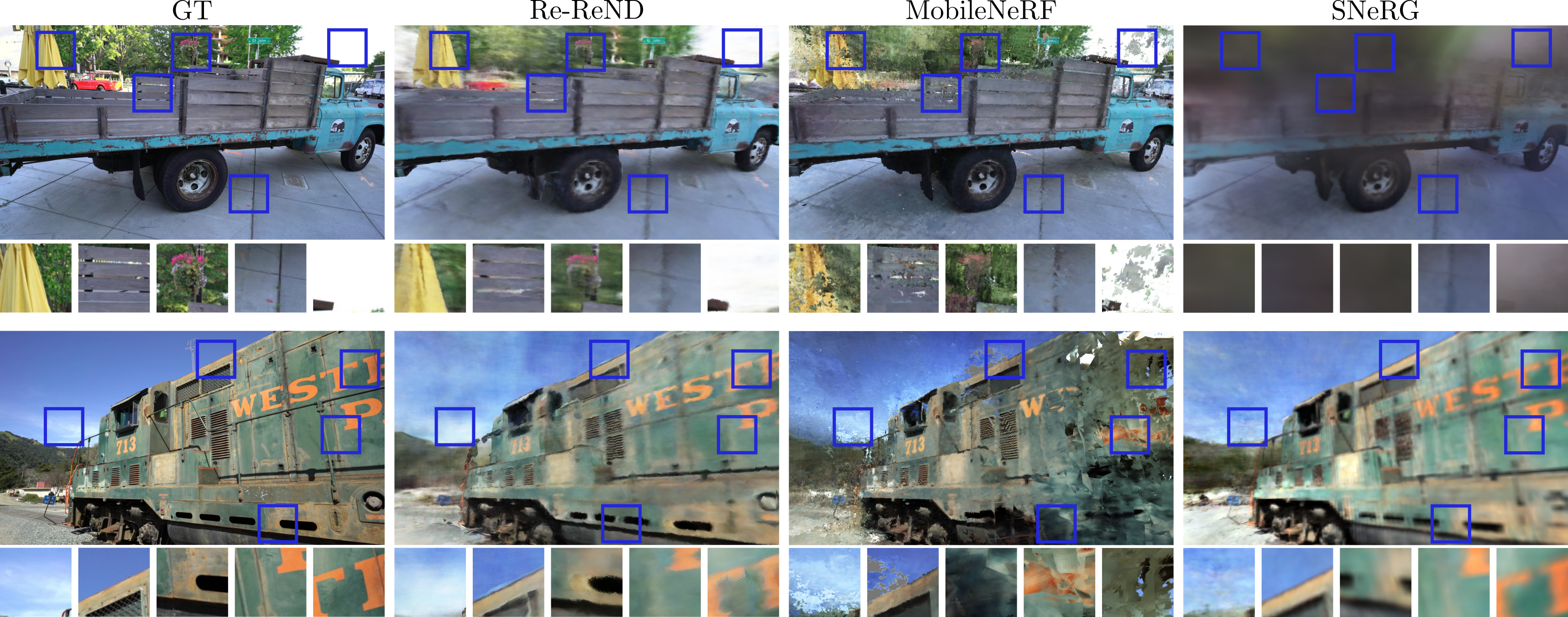}
    \caption{
    \textbf{Qualitative Results of Real Scenes.}
    We show renderings of challenging real scenes using \methodname, MobileNeRF~\cite{chen2022mobilenerf}, and SNeRG~\cite{hedman2021snerg}. 
    Our framework can generate sharper and more accurate renders. 
    In contrast, both MobileNeRF and SNeRG struggle with foreground and background artifacts, and fail to reconstruct fine details such as the writings on the train (bottom images). 
    }
    \vspace{-0.5cm}
    \label{fig:RealQualitative}
\end{figure*}

\subsection{Main Results}
We compare against two methods intended for fast rendering, SNeRG~\cite{hedman2021snerg} and MobileNeRF~\cite{chen2022mobilenerf}. 
\vspace{-0.5cm}
\paragraph{Quantitative Results.}
The first three rows of \Table{desktop} show how \methodname can achieve real-time rendering when deployed on a mobile.
For both datasets, we find that, while SNeRG achieves reasonable disk usage and photo-metric quality, it is ultimately incapable of interactive rendering.
\underline{On the Synthetic 360° dataset}, we find that \methodname can render at over 54 FPS, outperforming Mobile-NeRF by more than 30\%.
Furthermore, such fast rendering speed comes at a negligible drop in performance of $\sim$1 dB. 
Moreover, these benefits over MobileNeRF are accompanied by simpler meshes (\ie faces and vertex quantity).
\underline{On the realistic Unbounded 360° dataset}, our method outperforms competing approaches by even larger margins than on synthetic data.
In particular, we improve upon MobileNeRF \textit{both} in rendering quality and speed.
That is, while SNeRG and MobileNeRF achieve a PSNR of 14.0 and 15.6, respectively, \methodname attains 17.9.
Impressively, this outstanding gain in photo-metric quality of over 2 dB is accompanied by an also impressive superiority in rendering speed of over 10 FPS over MobileNeRF~(22.79 \textit{vs.} 33.46 FPS).
The gains we achieve can be attributed to our light field-based formulation, that allows us to work with smaller meshes while still maintaining high directional expressivity.
The bottom rows of \Table{desktop} report the performance of all methods in a non-constrained scenario, which is a desktop.
When provided with such larger computational resources, our method's speed improves both in absolute and relative terms.
Specifically, while the gap in mobiles w.r.t. MobileNeRF was $\sim$30\% (on the Synthetic dataset), this gap jumps to over 460\% on the desktop.
This increased gap is even more pronounced in the Unbounded dataset: the gap in mobiles was $\sim$45\% (from 22 FPS to 33), while the gap in desktop becomes over 1,500\%.
The larger computational resources also allow us to test a larger version of \methodname, which achieves remarkable PSNRs of 30.1 and 18.0 in the Synthetic and Unbounded datasets, respectively, while also remaining much faster than MobileNeRF.
The last three rows of \Table{desktop} report foundational methods (\ie NeRF, Mip-NeRF and NeRF++) as reference of high quality, although these methods are inefficient for rendering.
\vspace{-0.5cm}

\paragraph{Qualitative Results.} We next compare \methodname performance against the other real-time rendering methods. 
We show 
results for both synthetic (\Figure{SyntheticQualitative}) and real scenes (\Figure{RealQualitative}).
Note \methodname higher quality, particularly in real scenes, where our method preserves sharper object boundaries, and crisper surface details. 
\methodname allows real-time rendering while preserving strong image quality.

\begin{table}[t!]
    \footnotesize
    \centering
    \setlength{\tabcolsep}{1.5pt}
    \begin{tabular}{l"c|c|c"c|c|c}
        \thickhline
         & \multicolumn{3}{c"}{Synthetic 360°} & \multicolumn{3}{c}{Unbounded 360°} \\
        Device & SNeRG & M-NeRF & \methodname & SNeRG & M-NeRF & \methodname \\
        \thickhline
        Samsung S21 & $\dagger$ & \textcolor{black}{41.7} & \textcolor{black}{\best{54.7}} & $\dagger$ & \textcolor{black}{22.8} & \textcolor{black}{\best{33.5}} \\
        G9 & $\dagger$ & \textcolor{black}{9.7} & \textcolor{black}{\best{10.4}} & $\dagger$ & \textcolor{black}{3.5} & \textcolor{black}{\best{7.7}} \\
        \textcolor{black}{Galaxy S6 } & N/A & \textcolor{black}{18.1} & \textcolor{black}{\best{26.6}} & $\dagger$ & \textcolor{black}{6.2} & \textcolor{black}{\best{20.4}} \\
        Dell & $\dagger$ & \textcolor{black}{49.8} & \textcolor{black}{\best{75.3}} & $\dagger$ & \textcolor{black}{16.9} & \textcolor{black}{\best{54.0}} \\
        Gaming & \textcolor{black}{197.9}& \textcolor{black}{496.0} & \textcolor{black}{\best{697.3}} & \textcolor{black}{46.0} & \textcolor{black}{169.9} & \textcolor{black}{\best{516.6}} \\
        Desktop & \textcolor{black}{502.1} & \textcolor{black}{762.3} & \textcolor{black}{\best{1,013.2}} & \textcolor{black}{141.1} & \textcolor{black}{389.8} & \textcolor{black}{\best{925.4}} \\
        Headset & $\dagger$ & $-$ & \best{74.0}$^{\star}$ &  $\dagger$ & $-$ & \best{74.0}$^{\star}$ \\
        \thickhline
    \end{tabular}
    \vspace{0.07cm}
    \caption{\textbf{On-device rendering speed (FPS).} 
    We compare the rendering speed of \methodname against MobileNeRF (M-NeRF)~\cite{chen2022mobilenerf} and SNeRG~\cite{hedman2021snerg} across devices. 
    Conventions: $``^{\star}"$ means the device's FPS limit was reached, $``-"$ means missing implementation, and ``$\dagger$" means the method failed to run.
    For all devices and datasets, \methodname provides the fastest rendering speeds, usually by a large margin.
    Furthermore, we highlight that \methodname is capable of real-time rendering even on VR headsets. 
    }
    \vspace{-0.6cm}
    \label{tab:1_Rendering_speed_tb}
\end{table}

\subsection{Rendering Speed on Devices}
We measure frames-per-second (FPS) in seven devices~(via the same procedure as~\cite{chen2022mobilenerf}), and report results in \Table{1_Rendering_speed_tb}.
Our results illustrate how \methodname significantly outperforms other methods in rendering speed. 
The advantages in performance are particularly large in unbounded scenes. 
In this scenario, \methodname is, on average $2.6\times$ faster than MobileNeRF. 
For synthetic scenes, our method provides sizable speed gains of 35\%. 
Importantly, we find that \methodname is capable of real-time rendering in a VR headset even reaching the device's limit of 74 FPS. 

\subsection{Quality \textit{vs.} Representation Size}
\begin{figure}[t]
    \centering
    \includegraphics[width=1.0\linewidth]{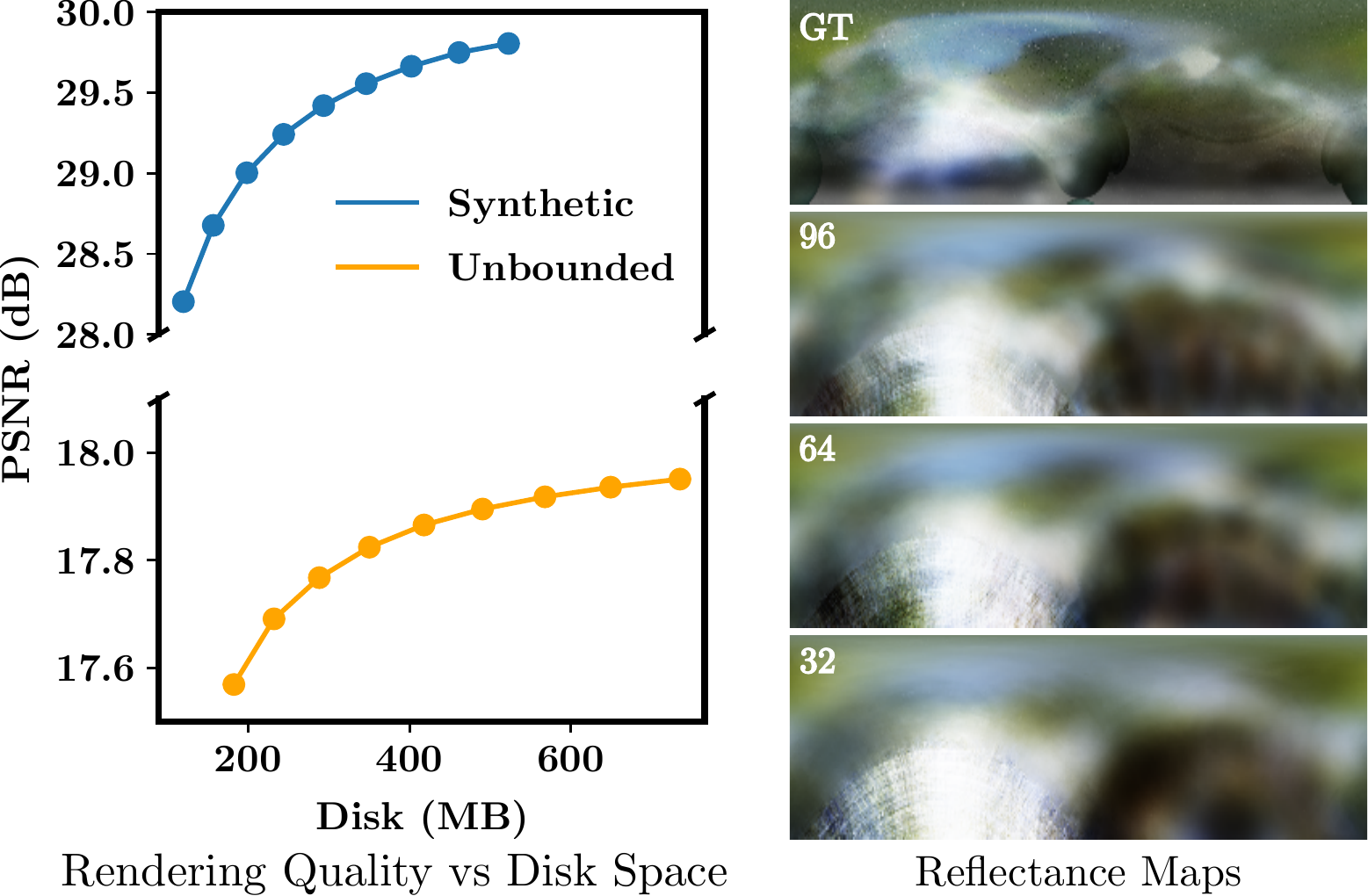}
    \caption{
    \textbf{Left:} We compare how PSNR varies as the number of texels per face increases for both synthetic and unbounded scenes.
    It is possible to trade-off disk space for higher quality renderings by increasing the number of texels.
    \textbf{Right:} We visualize color as a function of elevation (y-axis) and azimuth (x-axis) for a surface point x, y, z on the Materials scene from the Synthetic dataset, for different embedding dimensions. 
    }
    \vspace{-0.4cm}
    \label{fig:ablation}
\end{figure}

\begin{figure}[t]
    \includegraphics[width=\linewidth]{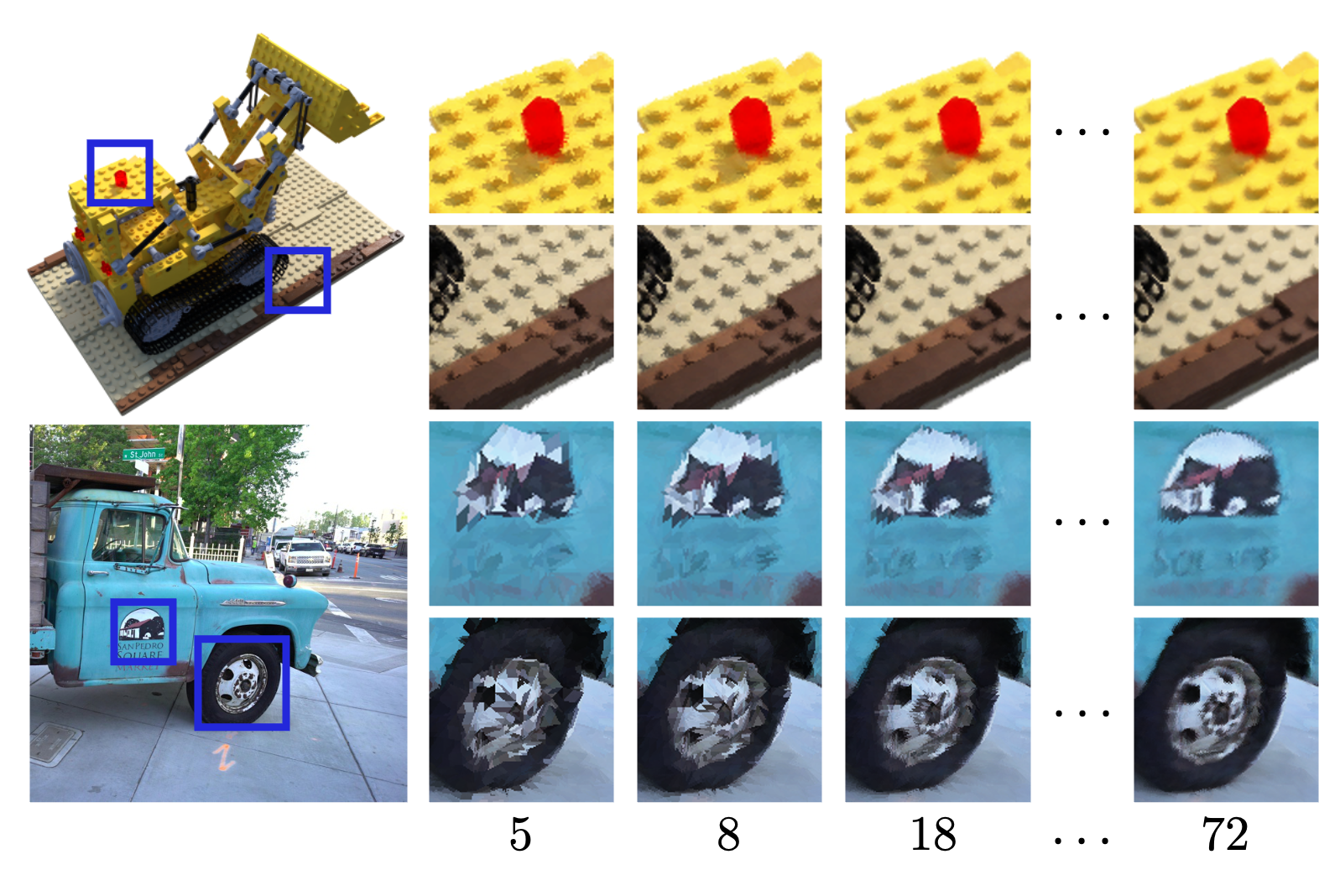}
    \vspace{-0.5cm}
    \caption{
    \textbf{Qualitative Effect of Varying Number of Texels.} 
    We show renderings of synthetic and real scenes with an increasing number of texels per triangle face.
    Higher texels per face result in crispier renderings, although it is possible to appreciate directional effects such as specular highlights even with low texel counts.
    }
    \vspace{-0.5cm}
    \label{fig:ablation2}
\end{figure}

Two main factors affect the size of our representation: the texel count and the dimensionality of the light field embeddings ($D$ in Eq.~\eqref{eq:dir-pos}).
Here we study how these factors, in turn, affect the photo-metric quality of \methodname.
\underline{Texel count.} 
We examine the effect that varying the number of texels assigned to each face in the mesh has on rendering quality and disk space. 
\Figure{ablation} (left) and \Figure{ablation2} show that, for both datasets, an increased texel count is accompanied by a drastic rise in quality of the reconstruction and disk space. 
\underline{Embedding dimensionality.} 
We now examine the effect that varying the dimensionality of the light field embeddings has on the ability to represent light effects.
Reflectance maps in \Figure{ablation} (right) demonstrate that \methodname can model challenging view-dependent effects even with low-dimensional embeddings, but larger embeddings enable more complex reflections. 
Together, our experiments demostrate that exchanging disk space for renderings of higher quality, and suggest that increasing the texel count can improve rendering quality.

\subsection{Compositional scenes}

\begin{figure}[t]
    \centering
    \includegraphics[width=\columnwidth]{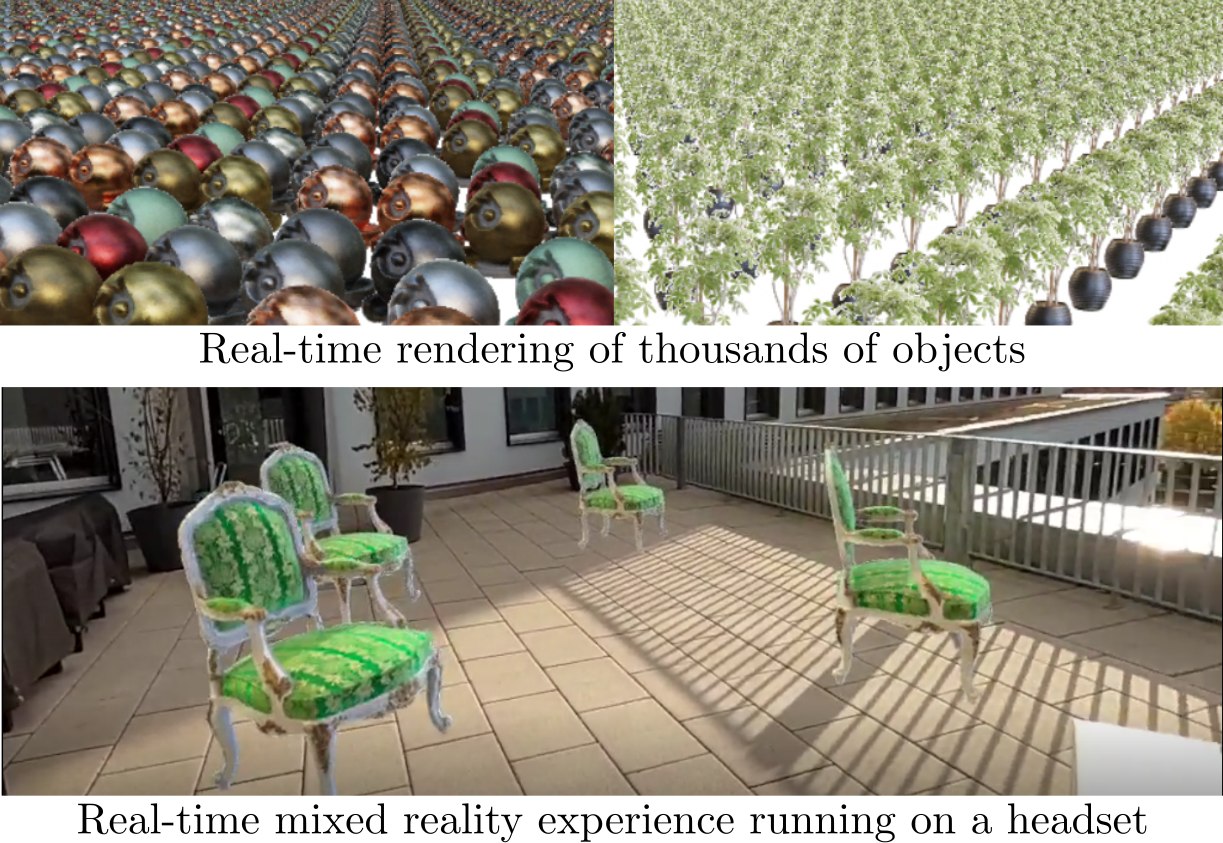}
    \caption{
    \textbf{Scene Composition.} \textbf{Top:} The outcome of rendering 2500 materials and ficus objects together in single scenes.  \textbf{Bottom:} An AR application showcase of \methodname with four chairs in a real-world setting using an AR/VR headset. See the \textbf{supplementary video} for more details.
    }
    \vspace{-0.5cm}
    \label{fig:app}
\end{figure}

In \Figure{app}, we showcase the practical application of \methodname for scene composition.
Our approach enables efficient rendering of 2500 materials and ficus objects in single scenes at 130 FPS each on a desktop.
Additionally, we demonstrate an AR application that uses an AR/VR headset to render four chairs in real-time in a real-world setting.
This exemplifies our approach's ability to seamlessly blend virtual and real environments, unlocking new possibilities for immersive visual experiences.


\textbf{Additional Results.} Finally, we report detailed qualitative and quantitative results, validation of view-dependent effects, sensitivity to geometry variations and photo-metric quality depending on the dimensionality $D$ of \methodname in the \SM.

\section{Conclusions and Limitations}\label{sec:conclusions}

We present \textbf{\methodname}, a method enabling \underline{Re}al-time \underline{Re}ndering of \underline{N}eRFs across \underline{D}evices. 
Our method receives a pre-trained NeRF as input and transforms it into a representation that renders real-time in various devices, from mobile phones to VR headsets.
Our approach achieves fast rendering by leveraging standard graphics pipelines, obtaining pixel color with a single query, and avoiding MLP evaluations altogether.
We achieve these objectives by distilling NeRF and extracting density into a mesh, and color information into matrices for efficient light field computation.

We test \methodname on a variety of resource-constrained devices, and find outstanding trade-off between rendering time and photo-metric quality. 
In particular, \methodname achieves over $2.6\times$ FPS improvements when rendering challenging and unbounded real scenes. 
Furthermore, our method allows rendering of these scenes on VR headsets in real-time, a task out of hand for current NeRF architectures. 

Limitations of our approach include relying on pre-trained NeRF for reasonable scene-geometry reconstructions and the requirement for large embeddings to achieve sensible reconstructions. Improvements in these areas could further enhance rendering speed and quality.

{\small
\bibliographystyle{ieee_fullname}
\bibliography{egbib}
}

\newpage
\section{Appendix}
\begin{figure}
    \centering
    \includegraphics[width=\columnwidth]{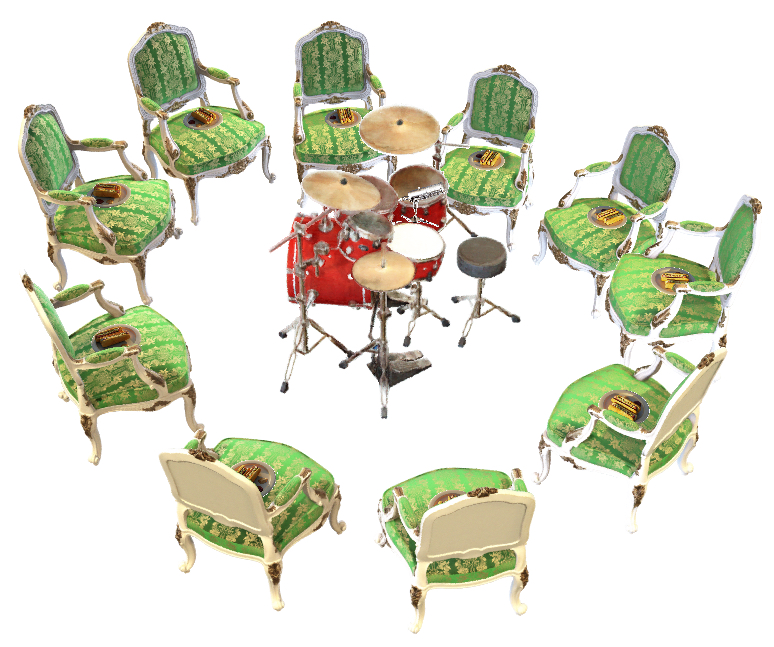}
    \caption{\textbf{\methodname enables real-time rendering of scenes that can be composed by tens of objects.} 
    Please refer to the video \texttt{Re-ReND.mp4}, demonstrating real-time rendering of this scene.
  }
    \label{fig:composit}
    \vspace{-.3cm}
\end{figure}

\begin{figure*}[t!]
    \centering
    \includegraphics[width=0.95\columnwidth]{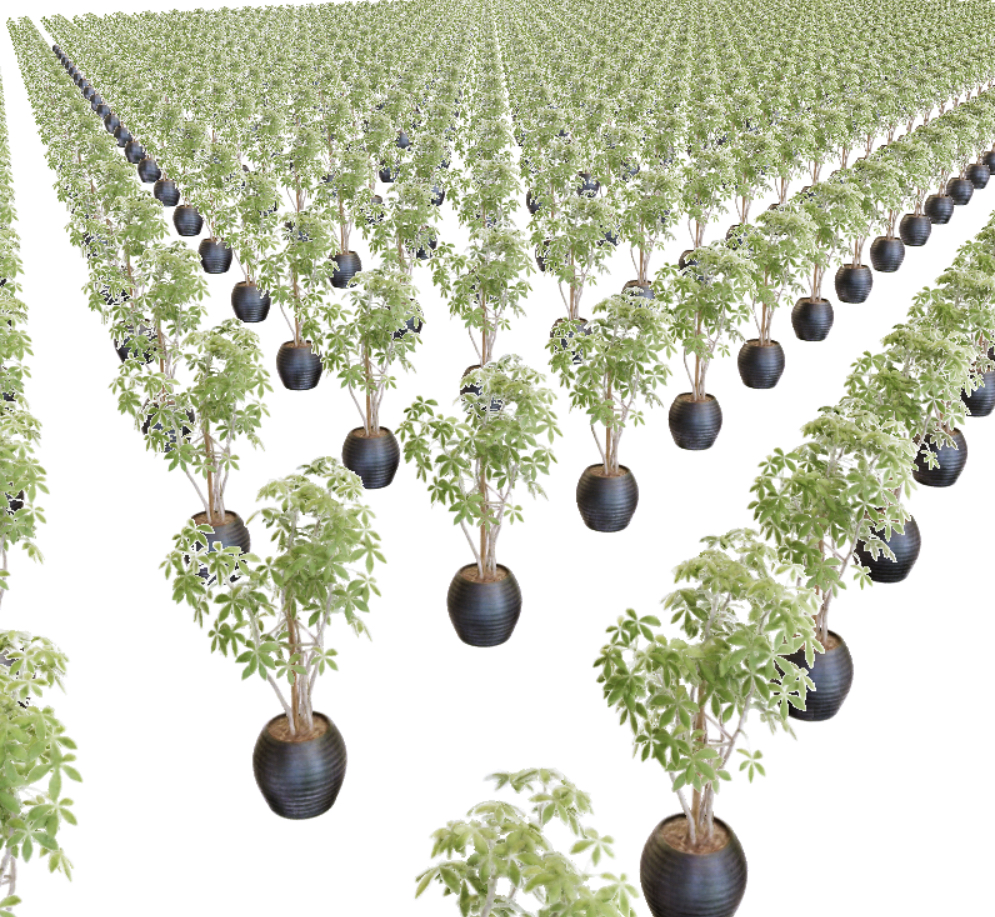}
    \hspace{0.5cm}
    \includegraphics[width=0.95\columnwidth]{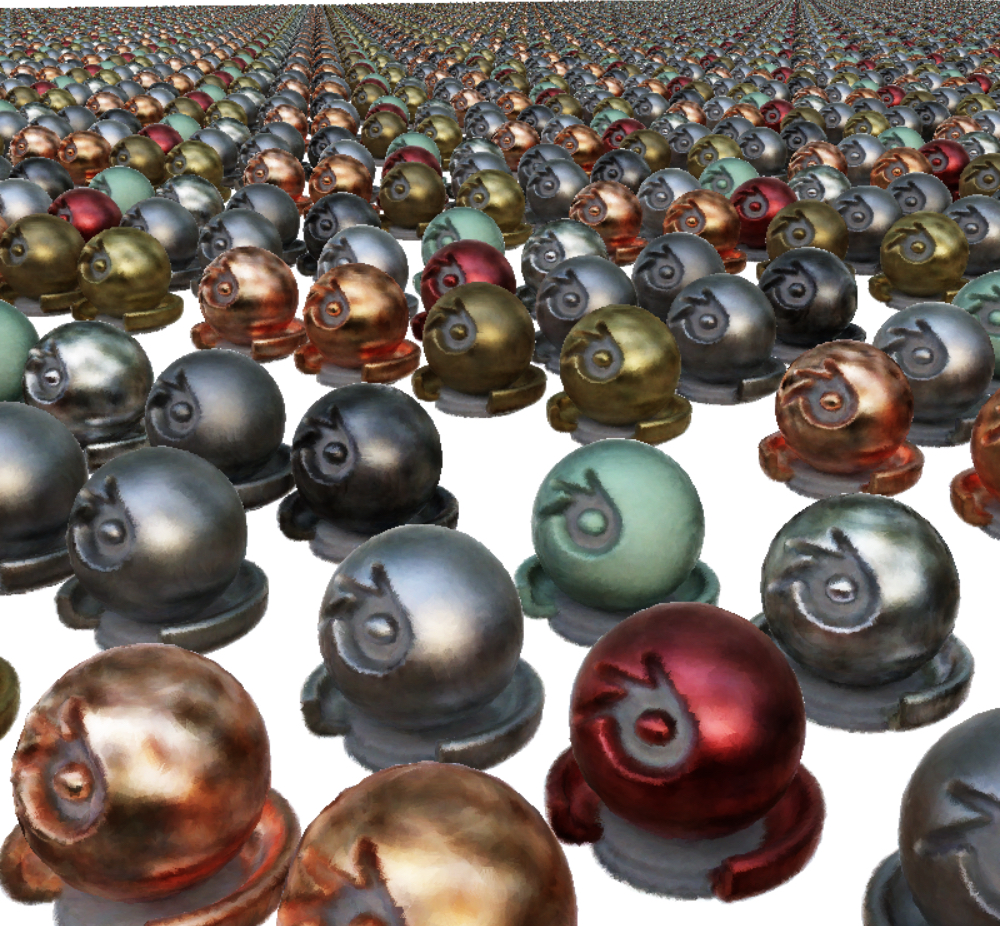}
    \vspace{0.2cm}
    \caption{\textbf{With \methodname, we can simultaneously render thousands of objects in real time.}
    Here we show 2,500 ficus scenes (left) and materials scenes (right).
    Please refer to the accompanying video \texttt{Re-ReND.mp4}, demonstrating real-time rendering of these scenes.
  }
  \vspace{-.3cm}
    \label{fig:many_objects}
\end{figure*}

Next, we present the Supplementary Materials for the paper ``Re-ReND: Real-time Rendering of NeRFs across Devices''.
Specifically, in addition to the results reported in the paper, we report results of \methodname w.r.t. Image Quality~(Section~\ref{sec:im_qual}) and (Section~\ref{sec:quali}), Rendering Speed~(Section~\ref{sec:fps}), Mesh Size~(Section~\ref{sec:mesh_size} and Section~\ref{sec:meshi}), Disk Space~(Section~\ref{sec:disk_space}), validation of view-dependent effects (Section~\ref{sec:val}),  sensitivity to geometry variations (Section~\ref{sec:geo}) and Photo-metric quality w.r.t. embedding dimensionality $D$ (Section~\ref{sec:dim}).
Furthermore, we encourage the reviewers to watch the \textbf{associated video}, \texttt{Re-ReND.mp4}, demonstrating \methodname's capabilities of real-time rendering across devices.
This video demonstrates how \methodname can render, in real time, a scene composed of tens (\Figure{composit}) or even thousands (\Figure{many_objects}) of objects. 
\Figure{composit} illustrates such a scene, composed of moving chairs, hotdogs, the drumset, and a microphone.








\section{Quantitative Results}
\subsection{Image Quality}\label{sec:im_qual}
At rendering, the image quality achieved by \methodname depends on the amount of texels,~\ie~pixels in the texture map, assigned to each triangle in the mesh.
Here, we report the effect that this variable has on image quality.
For assessing image quality, we measure the standard quality metrics~(PSNR, SSIM~\cite{SSIM} and LPIPS~\cite{LPIPS}) on both datasets~(\textit{Realistic Synthetic 360°} and \textit{360° Unbounded Tanks and Temples}).
Here we report disaggregate per-scene measures of PSNR (\Table{PSNRsyn} and \Table{PSNRreal}), SSIM (\Table{SSIMsyn} and \Table{SSIMreal}), and LPIPS (\Table{LPIPSsyn} and \Table{LPIPSreal}) for \methodname.

The traditional configuration of texels is equally-sized triangles in a texture map.
Thus, the number of texels (\ie the column ``$\:\text{Tex.}$'' in Tables~\ref{tab:PSNRsyn}-\ref{tab:LPIPSreal}) increases quadratically w.r.t. the triangle's side.
Formally, \#$\:\text{Tex.} = \lceil\nicefrac{p^2}{2}\rceil$, where $p$ is the number of pixels in the triangle's side.

Naturally, across all metrics and datasets, image quality improves as the number of texels increases.
We find that \methodname provides competitive performance when using $18$ texels per triangle.
However, while performance improves by increasing the number of texels, the quadratic growth of texels makes performance gains rapidly reach diminishing returns.


\begin{table}[ht]
\footnotesize
\centering
\setlength{\tabcolsep}{2pt}
\begin{tabular}{c"llllllll"l}\thickhline
 Tex. & chair & drums & ficus & hotd. & lego  & mat. & mic   & ship  & aver.     \\\thickhline
5       & 28.69 & 23.28 & 27.25 & 31.71  & 28.89 & 26.59  & 28.69 & 24.62 & 27.46   \\
8       & 29.66 & 23.75 & 27.57 & 32.55  & 30.13 & 27.16  & 29.78 & 25.04 & 28.20   \\
13      & 30.40 & 24.02 & 27.73 & 33.09  & 30.93 & 27.49  & 30.46 & 25.29 & 28.68   \\
18      & 30.99 & 24.19 & 27.83 & 33.48  & 31.48 & 27.71  & 30.91 & 25.45 & 29.00   \\
24      & 31.46 & 24.31 & 27.90 & 33.75  & 31.87 & 27.85  & 31.24 & 25.56 & 29.24   \\
32      & 31.85 & 24.39 & 27.94 & 33.95  & 32.17 & 27.95  & 31.48 & 25.63 & 29.42   \\
40      & 32.17 & 24.46 & 27.97 & 34.09  & 32.39 & 28.03  & 31.66 & 25.68 & 29.56   \\
50      & 32.43 & 24.50 & 27.99 & 34.21  & 32.57 & 28.08  & 31.81 & 25.72 & 29.66   \\
60      & 32.64 & 24.54 & 28.01 & 34.30  & 32.71 & 28.13  & 31.92 & 25.75 & 29.75   \\
72      & 32.83 & 24.56 & 28.02 & 34.37  & 32.82 & 28.16  & 31.91 & 25.78 & 29.81    \\\thickhline
\end{tabular}
\vspace{.1cm}
\caption{\textbf{PSNR on the Realistic Synthetic 360° dataset.} 
}
\vspace{-.3cm}
\label{tab:PSNRsyn}
\end{table}
\begin{table}
\footnotesize
\centering
\setlength{\tabcolsep}{4pt}
\begin{tabular}{c"llll"l}\thickhline
Tex. & truck & train & m60   & playg. & average \\\thickhline
5    & 14.92 & 19.31 & 16.70 & 18.53  & 17.37   \\
8    & 15.15 & 19.46 & 16.87 & 18.80  & 17.57   \\
13   & 15.27 & 19.56 & 16.98 & 18.96  & 17.69   \\
18   & 15.34 & 19.62 & 17.05 & 19.06  & 17.77   \\
24   & 15.40 & 19.66 & 17.10 & 19.14  & 17.82   \\
32   & 15.44 & 19.69 & 17.14 & 19.19  & 17.87   \\
40   & 15.47 & 19.71 & 17.17 & 19.23  & 17.90   \\
50   & 15.49 & 19.72 & 17.20 & 19.26  & 17.92   \\
60   & 15.51 & 19.74 & 17.22 & 19.29  & 17.94   \\
72   & 15.52 & 19.75 & 17.23 & 19.31  & 17.95 \\\thickhline
\end{tabular}
\vspace{.1cm}
\caption{\textbf{PSNR of 360° Unbounded Tanks and Temples dataset.}
}
\label{tab:PSNRreal}
\end{table}
\begin{table}[]
\footnotesize
\centering
\setlength{\tabcolsep}{1pt}
\begin{tabular}{c"llllllll"l}\thickhline
   Tex. & chair & drums & ficus & hotd. & lego  & mat. & mic   & ship  & aver.     \\\thickhline
5       & 0.93  & 0.9   & 0.943 & 0.948  & 0.933 & 0.917  & 0.965 & 0.807 & 0.918   \\
8       & 0.942 & 0.908 & 0.947 & 0.955  & 0.946 & 0.926  & 0.971 & 0.814 & 0.926   \\
13      & 0.95  & 0.912 & 0.949 & 0.959  & 0.953 & 0.93   & 0.975 & 0.818 & 0.931   \\
18      & 0.955 & 0.915 & 0.95  & 0.962  & 0.958 & 0.933  & 0.977 & 0.821 & 0.934   \\
24      & 0.959 & 0.917 & 0.951 & 0.964  & 0.961 & 0.935  & 0.978 & 0.822 & 0.936   \\
32      & 0.963 & 0.918 & 0.951 & 0.965  & 0.963 & 0.936  & 0.979 & 0.824 & 0.937   \\
40      & 0.965 & 0.919 & 0.952 & 0.966  & 0.964 & 0.937  & 0.98  & 0.824 & 0.938   \\
50      & 0.967 & 0.92  & 0.952 & 0.967  & 0.965 & 0.938  & 0.98  & 0.825 & 0.939   \\
60      & 0.968 & 0.92  & 0.952 & 0.967  & 0.966 & 0.938  & 0.98  & 0.825 & 0.940   \\
72      & 0.969 & 0.921 & 0.952 & 0.968  & 0.966 & 0.939  & 0.98  & 0.826 & 0.940  \\\thickhline
\end{tabular}
\vspace{.1cm}
\caption{\textbf{SSIM of Realistic Synthetic 360° dataset.} 
}
\vspace{-.2cm}
\label{tab:SSIMsyn}
\end{table}
\begin{table}[]
\footnotesize
\centering
\setlength{\tabcolsep}{4pt}
\begin{tabular}{c"llll"l}\thickhline
Tex. & truck & train & m60   & playg. & average \\\thickhline
5    & 0.486 & 0.513 & 0.456 & 0.535  & 0.498   \\
8    & 0.498 & 0.524 & 0.468 & 0.551  & 0.510   \\
13   & 0.506 & 0.533 & 0.477 & 0.563  & 0.520   \\
18   & 0.514 & 0.538 & 0.484 & 0.571  & 0.527   \\
24   & 0.521 & 0.543 & 0.489 & 0.578  & 0.533   \\
32   & 0.525 & 0.546 & 0.493 & 0.583  & 0.537   \\
40   & 0.529 & 0.548 & 0.496 & 0.586  & 0.540   \\
50   & 0.532 & 0.55  & 0.498 & 0.589  & 0.542   \\
60   & 0.535 & 0.551 & 0.5   & 0.591  & 0.544   \\
72   & 0.537 & 0.552 & 0.501 & 0.593  & 0.546   \\\thickhline
\end{tabular}
\vspace{.1cm}
\caption{\textbf{SSIM of 360° Unbounded Tanks and Temples dataset.} 
}
\vspace{-.2cm}
\label{tab:SSIMreal}
\end{table}
\begin{table}[]
\footnotesize
\centering
\setlength{\tabcolsep}{1.5pt}
\begin{tabular}{c"llllllll"l}\thickhline
   Tex. & chair & drums & ficus & hotd. & lego  & mat. & mic   & ship  & aver.     \\\thickhline
5       & 0.072 & 0.119 & 0.069 & 0.093  & 0.102 & 0.105  & 0.062 & 0.219 & 0.105   \\
8       & 0.062 & 0.109 & 0.061 & 0.079  & 0.083 & 0.094  & 0.052 & 0.209 & 0.094   \\
13      & 0.055 & 0.101 & 0.056 & 0.07   & 0.069 & 0.086  & 0.046 & 0.202 & 0.086   \\
18      & 0.049 & 0.094 & 0.053 & 0.064  & 0.06  & 0.08   & 0.041 & 0.197 & 0.080   \\
24      & 0.045 & 0.09  & 0.051 & 0.059  & 0.053 & 0.076  & 0.037 & 0.193 & 0.076   \\
32      & 0.042 & 0.086 & 0.049 & 0.056  & 0.049 & 0.073  & 0.035 & 0.191 & 0.073   \\
40      & 0.039 & 0.084 & 0.048 & 0.053  & 0.045 & 0.07   & 0.032 & 0.189 & 0.070   \\
50      & 0.037 & 0.082 & 0.047 & 0.052  & 0.042 & 0.068  & 0.031 & 0.187 & 0.068   \\
60      & 0.035 & 0.08  & 0.046 & 0.051  & 0.04  & 0.066  & 0.029 & 0.186 & 0.067   \\
72      & 0.034 & 0.079 & 0.046 & 0.05   & 0.038 & 0.065  & 0.028 & 0.185 & 0.066    \\\thickhline
\end{tabular}
\vspace{.1cm}
\caption{\textbf{LPIPS of Realistic Synthetic 360° dataset.} 
}
\vspace{-.2cm}
\label{tab:LPIPSsyn}
\end{table}
\begin{table}[]
\footnotesize
\centering
\setlength{\tabcolsep}{4pt}
\begin{tabular}{l"llll"l}\thickhline
Tex. & truck & train & m60   & playg. & average \\\thickhline
5    & 0.526 & 0.522 & 0.57  & 0.517  & 0.534   \\
8    & 0.519 & 0.514 & 0.565 & 0.507  & 0.526   \\
13   & 0.515 & 0.512 & 0.564 & 0.501  & 0.523   \\
18   & 0.513 & 0.51  & 0.563 & 0.496  & 0.521   \\
24   & 0.511 & 0.509 & 0.562 & 0.492  & 0.519   \\
32   & 0.509 & 0.507 & 0.56  & 0.49   & 0.517   \\
40   & 0.508 & 0.506 & 0.559 & 0.487  & 0.515   \\
50   & 0.506 & 0.505 & 0.557 & 0.485  & 0.513   \\
60   & 0.505 & 0.504 & 0.555 & 0.483  & 0.512   \\
72   & 0.504 & 0.504 & 0.553 & 0.481  & 0.51   \\\thickhline
\end{tabular}
\vspace{.1cm}
\caption{\textbf{LPIPS of 360° Unbounded Tanks and Temples dataset.} 
}
\vspace{-.2cm}
\label{tab:LPIPSreal}
\end{table}
\begin{table*}[ht]
\footnotesize
\centering
\setlength{\tabcolsep}{4pt}
\begin{tabular}{lllllllll|l||llll|l}\thickhline
            & chair  & drums  & ficus  & hotdog & lego   & materials & mic    & ship   & aver. & truck  & train  & playground & m60  & aver. \\\thickhline
Samsung S21                & 60.1   & 60.0   & 60.1  & 60.1   & 36.8  & 60.1      & 60.0   & 40.7  & 54.7    & 23.2  & 37.0  & 27.9       & 45.7  & 33.5    \\
Motorola G9                & 11.7   & 13.4   & 6.5   & 12.3   & 4.1   & 14.2      & 18.0   & 3.2   & 10.4    & 6.7   & 9.0   & 5.8        & 9.4   & 7.7     \\
Galaxy S6                  & 31.6   & 33.8   & 21.1  & 27.2   & 12.5  & 32.3      & 42.5   & 12.2  & 26.6    & 18.1  & 20.7  & 19.6       & 23.3  & 20.4    \\
Dell                       & 84.6   & 84.5   & 74.8  & 72.4   & 50.3  & 82.3      & 110.4  & 42.8  & 75.3    & 49.8  & 57.1  & 54.4       & 54.8  & 54.0    \\
Gaming                     & 769.2  & 762.7  & 688.3 & 684.4  & 447.2 & 759.5     & 1065.1 & 401.8 & 697.3   & 469.4 & 560.8 & 483.2      & 553.0 & 516.6   \\
PC                         & 1113.1 & 1130.7 & 997.0 & 1067.6 & 807.6 & 998.9     & 1117.0 & 873.4 & 1013.2  & 884.3 & 967.0 & 952.1      & 898.2 & 925.4 \\\thickhline
\end{tabular}
\caption{\textbf{Frames per second (FPS) achieved by \methodname.}
We report the disaggregated FPS for all devices we tested on all the scenes.
Columns 2-10: Realistic Synthetic 360° dataset.
Columns 11-15: 360° Unbounded Tanks and Temples dataset.
}
\vspace{-.2cm}
\label{tab:disaggregated_fps}
\end{table*}

\begin{table*}[ht]
\footnotesize
\centering
\setlength{\tabcolsep}{4pt}
\begin{tabular}{lllllllll|l||llll|l}\thickhline
         & chair  & drums  & ficus  & hotdog & lego   & materials & mic    & ship   & average  & truck  & train  & playground & tank   & average \\\thickhline
Faces    & 158k  & 164k  & 239k  & 146k   & 360k & 159k      & 131k & 284k & 205,693 & 265k  & 232k  & 246k       & 235k & 244,847 \\
Vertices & 76k   & 80k   & 119k  & 72k    & 171k & 76k       & 62k  & 136k & 99,539  & 125k  & 115k  & 116k       & 115k & 117,751 \\\thickhline
\end{tabular}
\caption{\textbf{Mesh sizes used by \methodname.}
We report the number of triangle faces and vertices used to model each scene.
Columns 2-10: Realistic Synthetic 360° dataset.
Columns 11-15: 360° Unbounded Tanks and Temples dataset.
}
\label{tab:mesh}
\end{table*}

\begin{table}[]
\footnotesize
\centering
\setlength{\tabcolsep}{1.5pt}
\begin{tabular}{l"llllllll"l}\thickhline
\# Tex. & chair & drums & ficus & hotd. & lego  & mater. & mic   & ship  & average \\\thickhline
5    & 69.1  & 75.7  & 107.3 & 66.4  & 141.7 & 73.4   & 60.5  & 111.3 & 88.1    \\
8    & 94.8  & 103.4 & 147.7 & 88.7  & 195.6 & 98.7   & 80.9  & 149.4 & 119.8   \\
13   & 126.0 & 135.7 & 193.9 & 114.4 & 259.1 & 128.0  & 105.1 & 193.3 & 156.9   \\
18   & 161.8 & 172.2 & 245.9 & 143.0 & 330.1 & 160.8  & 132.3 & 241.8 & 198.4   \\
24   & 201.7 & 212.0 & 303.0 & 174.0 & 408.3 & 196.7  & 162.6 & 294.7 & 244.1   \\
32   & 245.8 & 255.7 & 364.5 & 207.8 & 492.7 & 235.5  & 195.8 & 351.1 & 293.6   \\
40   & 293.6 & 302.1 & 430.0 & 243.5 & 583.0 & 276.9  & 231.2 & 411.2 & 346.4   \\
50   & 344.7 & 351.5 & 499.2 & 281.2 & 678.8 & 320.6  & 269.2 & 474.2 & 402.4   \\
60   & 399.2 & 403.6 & 571.7 & 320.6 & 779.6 & 366.5  & 309.4 & 540.3 & 461.3   \\
72   & 456.9 & 458.0 & 647.3 & 361.8 & 885.0 & 414.6  & 349.7 & 609.1 & 522.8     \\\thickhline
\end{tabular}
\caption{\textbf{Disk Space (MB) of Realistic Synthetic 360° dataset.}
}
\label{tab:Disksyn}
\end{table}
\begin{table}[]
\footnotesize
\centering
\setlength{\tabcolsep}{4pt}
\begin{tabular}{l"llll"l}\thickhline
\# Tex. & truck & train & m60   & playg. & average \\\thickhline
5    & 143.8 & 132.3 & 147.6 & 137.7  & 140.37  \\
8    & 185.3 & 172.3 & 192.9 & 179.1  & 182.42  \\
13   & 234.0 & 220.1 & 246.3 & 228.3  & 232.17  \\
18   & 288.7 & 273.6 & 307.0 & 283.9  & 288.31  \\
24   & 348.3 & 333.1 & 374.4 & 345.7  & 350.37  \\
32   & 413.3 & 398.1 & 448.2 & 412.5  & 418.04  \\
40   & 482.1 & 468.0 & 527.6 & 484.3  & 490.49  \\
50   & 555.1 & 542.3 & 612.7 & 561.5  & 567.89  \\
60   & 631.4 & 620.7 & 702.7 & 642.3  & 649.27  \\
72   & 712.7 & 702.6 & 798.1 & 727.4  & 735.18   \\\thickhline
\end{tabular}
\caption{\textbf{Disk Space (MB) of 360° Unbounded Tanks and Temples dataset.} 
}
\label{tab:Diskreal}
\end{table}

\subsection{Rendering Speed} \label{sec:fps}

We report the disaggregated rendering speeds achieved by \methodname, measured in frames per second (FPS), in \Table{disaggregated_fps}.
For the Realistic Synthetic 360° dataset, \methodname attains an average speed of over 54 FPS even on a Samsung~S21.


\subsection{Mesh Size} \label{sec:mesh_size}
Table~4 in the main paper reports the average size (number of vertices and triangle faces) of the meshes used by \methodname.
We report the per-scene mesh sizes in \Table{mesh} for both datasets we experimented with. 
Overall, for the Realistic Synthetic 360° dataset (left columns), \methodname uses, on average, fewer than 205k faces and 99k vertices.
For the 360° Unbounded Tanks and Temples dataset, these numbers correspond to 250k faces and 120k vertices.
As such, these meshes are decidedly not particularly precise, and thus serve mostly as a collision mesh for \methodname to estimate where the scene's geometry is.

\subsection{Disk Space}\label{sec:disk_space}
The number of texels assigned to each triangle in the mesh affects the disk space used for representing a scene.
We vary the number of texels, and report the disk space used for each scene in \Table{Disksyn}~(for Realistic Synthetic 360°) and \Table{Diskreal}~(for 360° Unbounded Tanks and Temples).

For the Realistic Synthetic 360° dataset, \methodname's default of $18$ texels implies using an average disk space of $198.8$ MB.
Furthermore, all the objects (except ficus, lego and ship), use fewer than $200$ MB.
On the other hand, for the 360° Unbounded Tanks and Temples dataset, the default of $18$ texels makes all scenes use a disk space between $270$ and $310$ MB.

\begin{figure}[ht]
    \centering
    \includegraphics[width=\columnwidth]{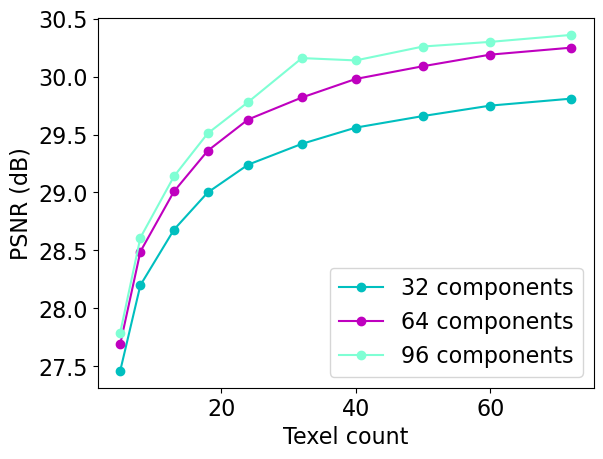}
    \caption{\textbf{Photo-metric quality depending on the dimensionality} 
   We report \methodname's results for various dimensionalities of embedding.
  }
    \label{fig:dim}
    \vspace{-.3cm}
\end{figure}
\section{Qualitative Results}

\subsection{Image Quality}\label{sec:quali}
\begin{figure*}[t]
    \centering
    \includegraphics[ width=1\textwidth]{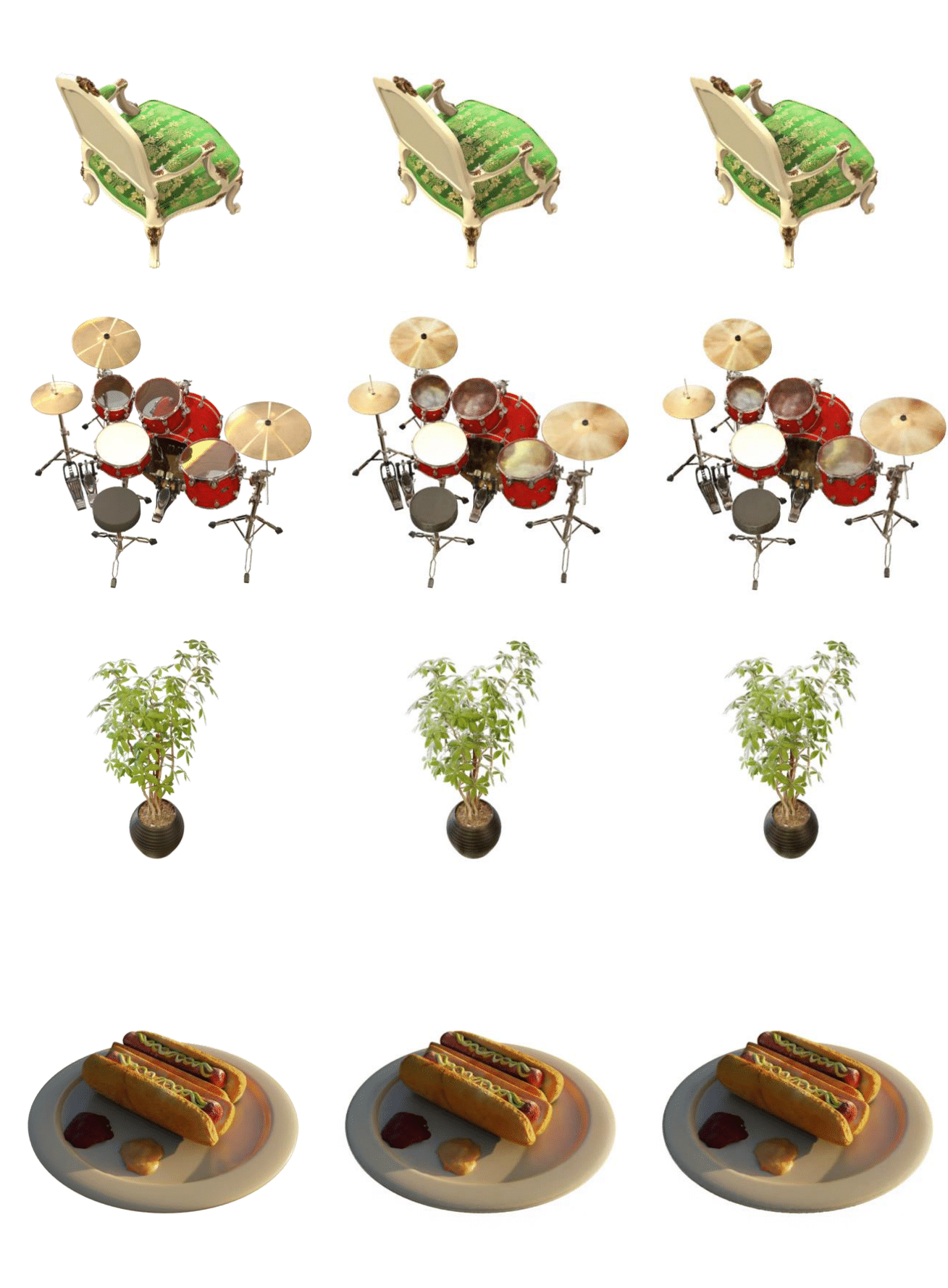}
    \caption{\textbf{Qualitative results.} We report the qualitative results for synthetic dataset. From left to right: first column is GT, second column is  \methodname using quad size 72 and 32 components, and third  columns is \methodname using quad size 18 and 32 components.
  }
    \label{fig:qlsyn}
\end{figure*}

\begin{figure*}[t]
    \centering
    \includegraphics[ width=1\textwidth]{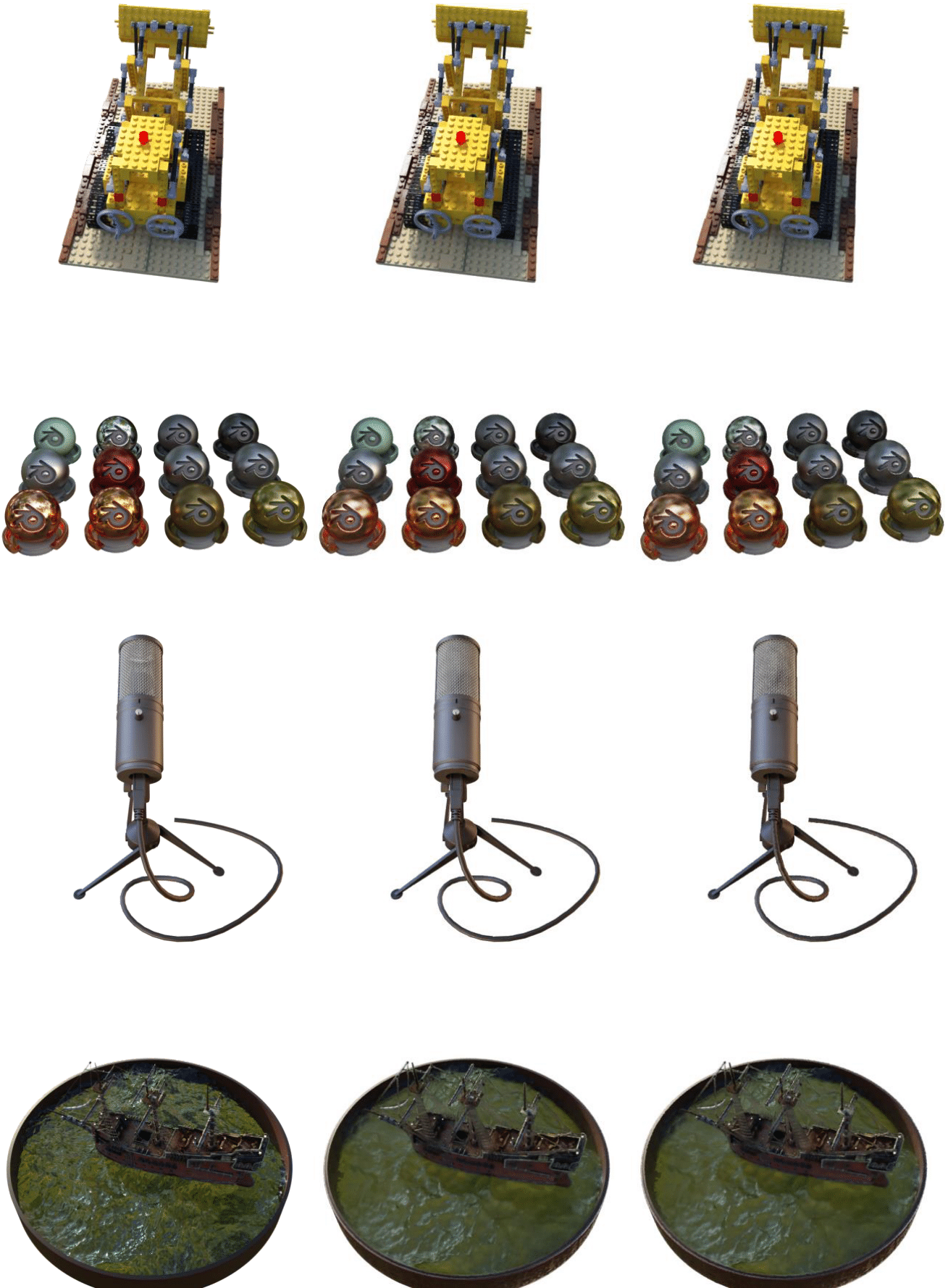}
    \caption{\textbf{Qualitative results.} We report the qualitative results for synthetic dataset. From left to right: first column is GT, second column is  \methodname using quad size 72 and 32 components, and third  columns is \methodname using quad size 18 and 32 components.
  }
    \label{fig:qlsyn1}
\end{figure*}

\begin{figure*}[t]
    \centering
    \includegraphics[ width=1\textwidth]{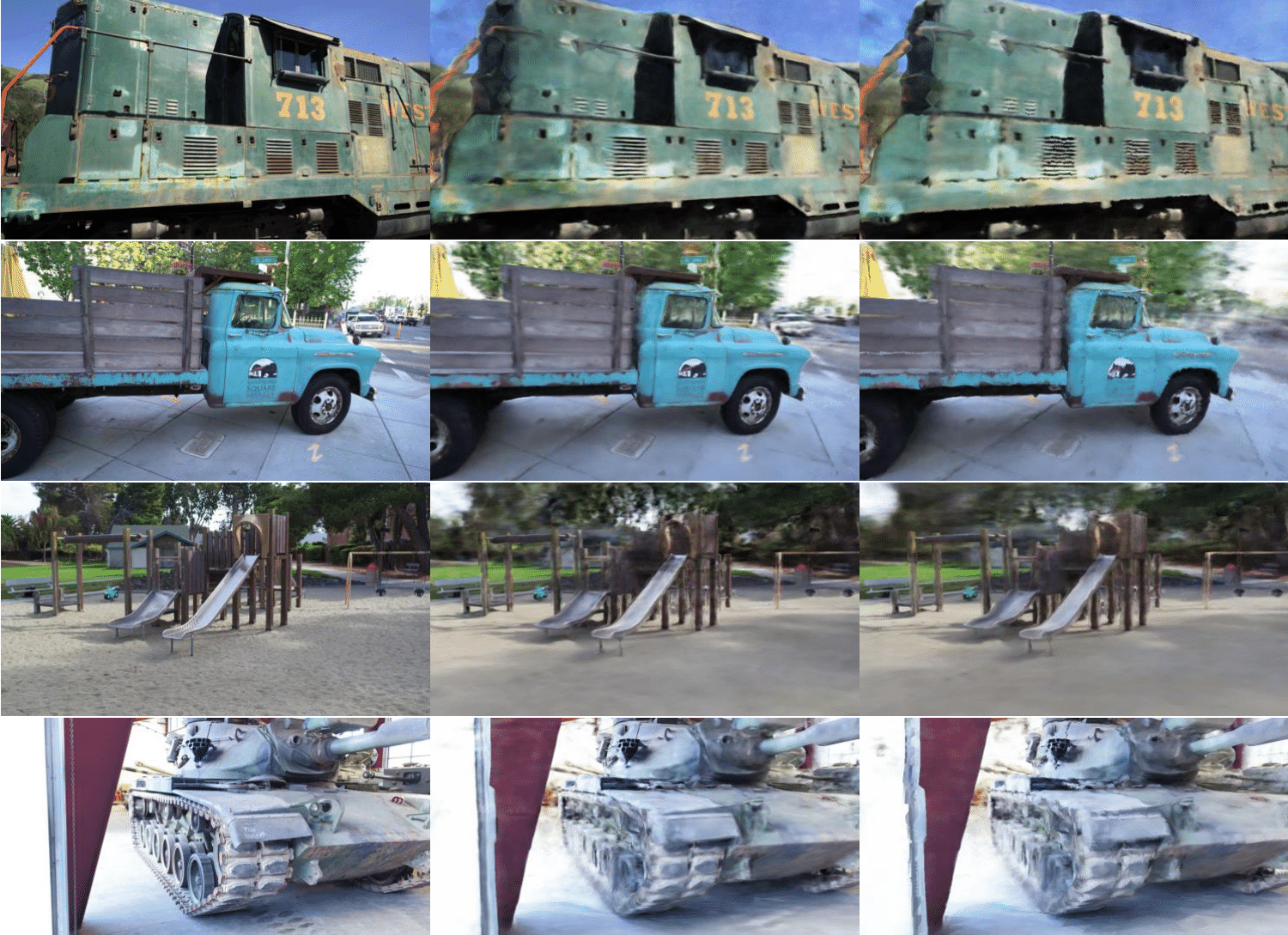}
    \caption{\textbf{Qualitative results.} We report the qualitative results for real 360 dataset. From left to right: first column is GT, second column is  \methodname using before discretization, and third columns is \methodname using quad size 18 and 32 components.
  }
    \label{fig:qlsyn2}
\end{figure*}
In \Figure{qlsyn}, and \Figure{qlsyn1}, we present the qualitative results obtained on a synthetic dataset. The images in the first column represent the ground truth (GT) data. In the second column, we show the results obtained using Re-ReND with quad size 72 and 32 components. Finally, in the third column, we display the results obtained using Re-ReND with quad size 18 and 32 components.

Upon visual inspection, we observe that Re-ReND with quad size 72 and 32 components produces more accurate and visually appealing results compared to Re-ReND with quad size 18 and 32 components. The former shows greater detail and smoother transitions between the different regions of the scene. However, Re-ReND with quad size 18 and 32 components still manages to produce decent results.

In addition to the synthetic dataset, we also present qualitative results on a real 360 dataset [See \Figure{qlsyn2}]. The images in the first column represent the ground truth (GT) data. In the second column, Re-ReND before discretization, while in the third column, Re-ReND with quad size 18 and 32 components.

It is worth noting that these qualitative results are obtained on a real 360 dataset, which presents more challenges compared to the synthetic dataset. The real-world scenario involves more complex lighting conditions, occlusions, and variations in scene geometry. 

\subsection{Meshes}\label{sec:meshes}\label{sec:meshi}
\begin{figure*}[t]
    \centering
    \includegraphics[width=0.7\textwidth]{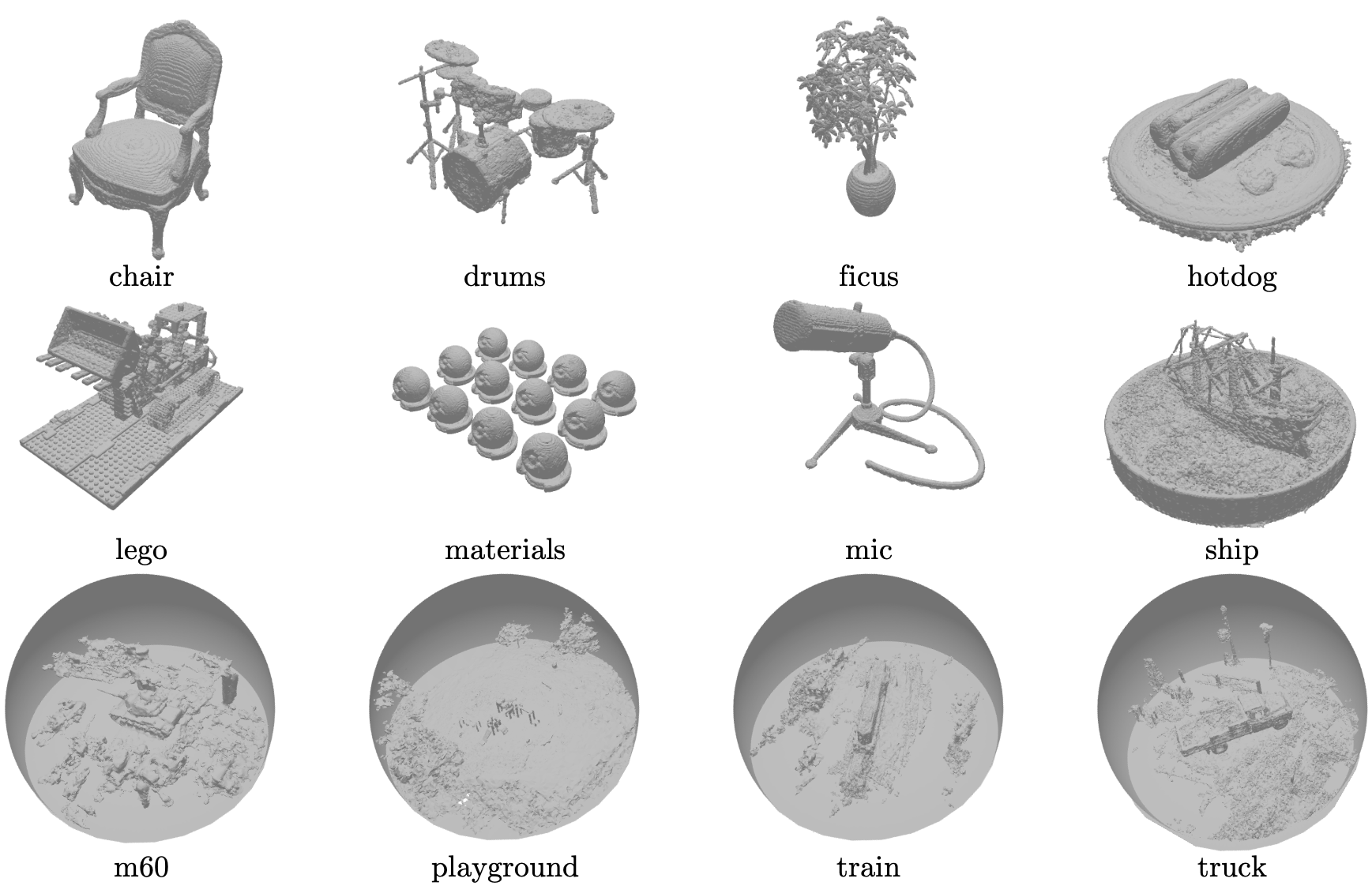}
    \caption{\textbf{Meshes used by \methodname.} 
  }
    \label{fig:meshes}
\end{figure*}
In \Figure{meshes}, we report the meshes we use in \methodname for both datasets.
All our meshes are simple and smooth.
Note that, for the 360° Unbounded Tanks and Temples dataset (last row in \Figure{meshes}), the scene is encapsulated within a semi-sphere and a plane mimicking the floor.

 \section{Validation of view-dependent effects}\label{sec:val}
 
 To validate the effect of view direction, we conducted an experiment comparing the performance of Re-ReND to a simple RGB textured mesh representation as a baseline. Due to the lack of ground truth RGB textures, we created a texture by assigning colors based on the intersected face's normal of a pretrained \methodname. The results showed that on both the Synthetic and Unbounded T\&T datasets, the RGB textured mesh PSNR was lower compared to Re-ReND. Specifically, the PSNR values were 22.82 dB and 14.79 dB for the RGB textured mesh representation, compared to Re-ReND's 29.00 dB and 17.77 dB, respectively.

This significant performance drop highlights the critical importance of modeling view-dependent effects for achieving high-quality image reconstruction. The Re-ReND approach, which takes into account view-dependent effects, was able to produce more accurate and visually appealing results than the simple RGB textured mesh representation. This finding suggests that the Re-ReND method is effective in modeling view-dependent effects and can lead to improved image reconstruction results.

 \section{Sensitivity to geometry variations}\label{sec:geo}
We conduct an experiment to evaluate the sensitivity of a Re-ReND to geometry quality. To do so, we use the ground truth meshes of the synthetic dataset to train Re-ReND, and we compared two sets of results: one using perfect geometry, and another using "cheap" meshes by marching cubes.

The results were then evaluated using three metrics before discretization: PSNR, SSIM, and LPIPS. The results for the perfect geometry case were 31.10, 0.954, and 0.0535 for PSNR, SSIM, and LPIPS, respectively. For the cheap mesh case, the results were 30.73, 0.946, and 0.0562 for the same metrics.

Analyzing the results, it appears that the Re-ReND is sensitive to geometry quality, as the results for the perfect geometry case were consistently better across all three metrics. This suggests that the method performs better when it has access to high-quality geometry information. However, even when using cheaper meshes, the Re-ReND method can still perform reasonably well. For example, the PSNR values were only slightly different between the perfect and cheap mesh cases, and the difference in LPIPS and SSIM values was within a reasonable range. This suggests that the Re-ReND method can perform very well without losing too much quality, even when the geometry information is not perfect.

 \section{Photo-metric quality depending on the dimensionality D}\label{sec:dim}
In \Figure{dim}, we report Re-ReND’s results for various dimensionalities of embedding. The dimensionality of the embedding is an important factor that can affect the performance of the model in various ways. On one hand, a higher dimensional embedding can potentially capture more complex textures and materials, leading to better PSNR. On the other hand, a higher dimensional embedding may also require more memory usage, making it slower and not apt for certain devices.

In practice, the choice of embedding dimensionality is often a trade-off between quality and efficiency, and depends on the specific requirements and constraints of the application. For example, in low-constraint devices, a lower dimensional embedding may be sufficient to achieve good performance, while for desktop, a higher dimensional embedding may be necessary to obtain better results in 8K resolution.

\end{document}